\documentclass[11pt]{article}

\usepackage[final]{acl}

\usepackage{times}
\usepackage{latexsym}
\usepackage{booktabs} 
\usepackage{enumitem}
\usepackage{makecell}
\usepackage{amsmath} 
\usepackage{amsfonts}
\usepackage{amsthm}
\newtheorem{theorem}{Theorem}
\newtheorem{lemma}{Lemma}[section]
\usepackage{pifont}
\newcommand{\cmark}{\ding{51}}
\newcommand{\xmark}{\ding{55}}

\usepackage{tabularx}
\usepackage{amssymb}

\usepackage{multirow}
\usepackage{graphicx}
\usepackage{adjustbox}

\usepackage{xcolor}       % \textcolor 和 blue!70!black 混色
\usepackage{fontawesome5} % \faGithub 图标
\usepackage{hyperref}     % \href 和 \hypersetup，通常靠后加载

\usepackage[table]{xcolor}
\definecolor{srmrgreen}{RGB}{0,128,64} % SRMR material retained for later removal

\usepackage[T1]{fontenc}
\usepackage[utf8]{inputenc}

\usepackage{microtype}

\usepackage{inconsolata}

\usepackage{graphicx}

\title{HOLogicBench: Evaluating Higher-Order Logical Reasoning in LLMs}

\author{%
\textbf{Yucheng Wu}$^{1,2}$\thanks{Equal contribution.} \quad
\textbf{Jundong Xu}$^{5}$\footnotemark[1] \quad
\textbf{Mingzhen Ju}$^{2}$ \quad
\textbf{Yue Yu}$^{2}$ \\
\textbf{Chenpeng Wang}$^{6}$ \quad
\textbf{Haoxuan Li}$^{3}$ \quad
\textbf{Liangming Pan}$^{1,2,4}$\thanks{Corresponding author.} \quad\\
\normalfont
$^1$State Key Laboratory of Multimedia Information Processing, Peking University \\
$^2$School of Computer Science, Peking University \\
$^3$Institute for Artificial Intelligence, Peking University \\
$^4$Beijing Academy of Artificial Intelligence, Beijing, China \\
$^5$School of Computing, National University of Singapore \\
$^6$YiXin-AILab, YIXIN, Beijing, China
}

\begin{document}
\maketitle

\begin{abstract}
Existing LLM logical-reasoning benchmarks largely focus on object-level inference under fixed predicates and rules, leaving higher-order reasoning over rules and policies underexplored.
We introduce \textsc{HOLogicBench}, a 1,379-instance benchmark that systematically evaluates five representative higher-order reasoning operations, with Isabelle/HOL formalizations and proof-verified answers.
Beyond answer accuracy, \textsc{HOLogicBench} provides solver-verified reference reasoning paths for each instance, enabling fine-grained evaluation of model-generated reasoning against valid derivations.
Across 16 model configurations, average accuracy is 53.86\%, with the best configuration reaching 65.99\%.
Further analyses reveal that LLMs exhibit limited alignment with valid derivations, remain vulnerable to minimal counterfactual perturbations, and display distinct failure patterns across higher-order reasoning types.
\end{abstract}

% \vspace{-0.5em}
\begin{center}
% \small
\begingroup
\hypersetup{pdfborder={0 0 0}}
\href{https://github.com/wuyucheng2002/HOLogicBench}
{\textcolor{black}{\faGithub}\hspace{0.6em}%
 \textcolor{blue!70!black}{\texttt{HOLogicBench}}}
\endgroup
\end{center}
% \vspace{-0.5em}

% \begin{abstract}
% Logical reasoning is essential for reliable AI systems, as it requires models to follow explicit rules and produce reasoning processes that are controllable, auditable, and verifiable. 
% Although recent benchmarks have advanced the evaluation of such abilities in large language models (LLMs), they are predominantly first-order-logic-centric, mainly testing object-level deduction over fixed predicates. 
% This leaves a critical gap: many realistic tasks require reasoning over rules, predicates, functions, constraints, and decision procedures themselves. 
% To address this gap, we introduce \textsc{HOLogicBench} (\textbf{H}igher-\textbf{O}rder \textbf{L}ogic \textbf{M}eets real-world \textbf{E}xplainable \textbf{S}ymbolic reasoning), the first real-world benchmark centered on higher-order reasoning for LLMs. 
% Built on higher-order logic, \textsc{HOLogicBench} provides natural-language problems, HOL formalizations, ground-truth answers, verifiable reasoning paths, and fine-grained skill annotations, enabling diagnostic evaluation of rule-level and function-level reasoning. 
% Experiments show that current LLMs struggle on faithful high-order reasoning, revealing failures in rule comparison, constraint composition, predicate/function-level reasoning, and formal reasoning trace generation. 
% These findings suggest that higher-order symbolic reasoning remains a major bottleneck for building reliable and verifiable LLMs.
% \end{abstract}

\section{Introduction}

Large language models (LLMs) have demonstrated increasingly strong capabilities on complex real-world tasks, including interactive decision-making and agentic problem solving~\citep{zhou2024webarena,mialon2024gaia}. 
As these systems are used in more consequential settings, logical reasoning becomes essential: models must follow explicit rules, combine conditions, resolve constraints, and justify decisions through verifiable reasoning steps. 
This capability is especially important in high-stakes domains such as law, medicine, scientific discovery, program verification, and financial compliance, where decisions must be logically sound, verifiable, and reliable.
Evaluating faithful logical reasoning is therefore crucial for assessing LLM reliability beyond task success \citep{cheng2025empowering,liu2025logical}.

% \begin{figure}[t]
%     \centering
%     \includegraphics[width=0.95\linewidth]{fig/intro.pdf}
%     \caption{
%     Illustration of the difference between first-order and higher-order reasoning. 
%     FOL reasoning typically evaluates deduction over objects under fixed predicates, whereas HOL reasoning can treat predicates, functions, and rules themselves as reasoning objects.
%     }
%     % \vspace{-1em}
%     \label{fig:fol-hol-comparison}
% \end{figure}

\begin{figure}[t]
    \centering
    \includegraphics[width=\linewidth]{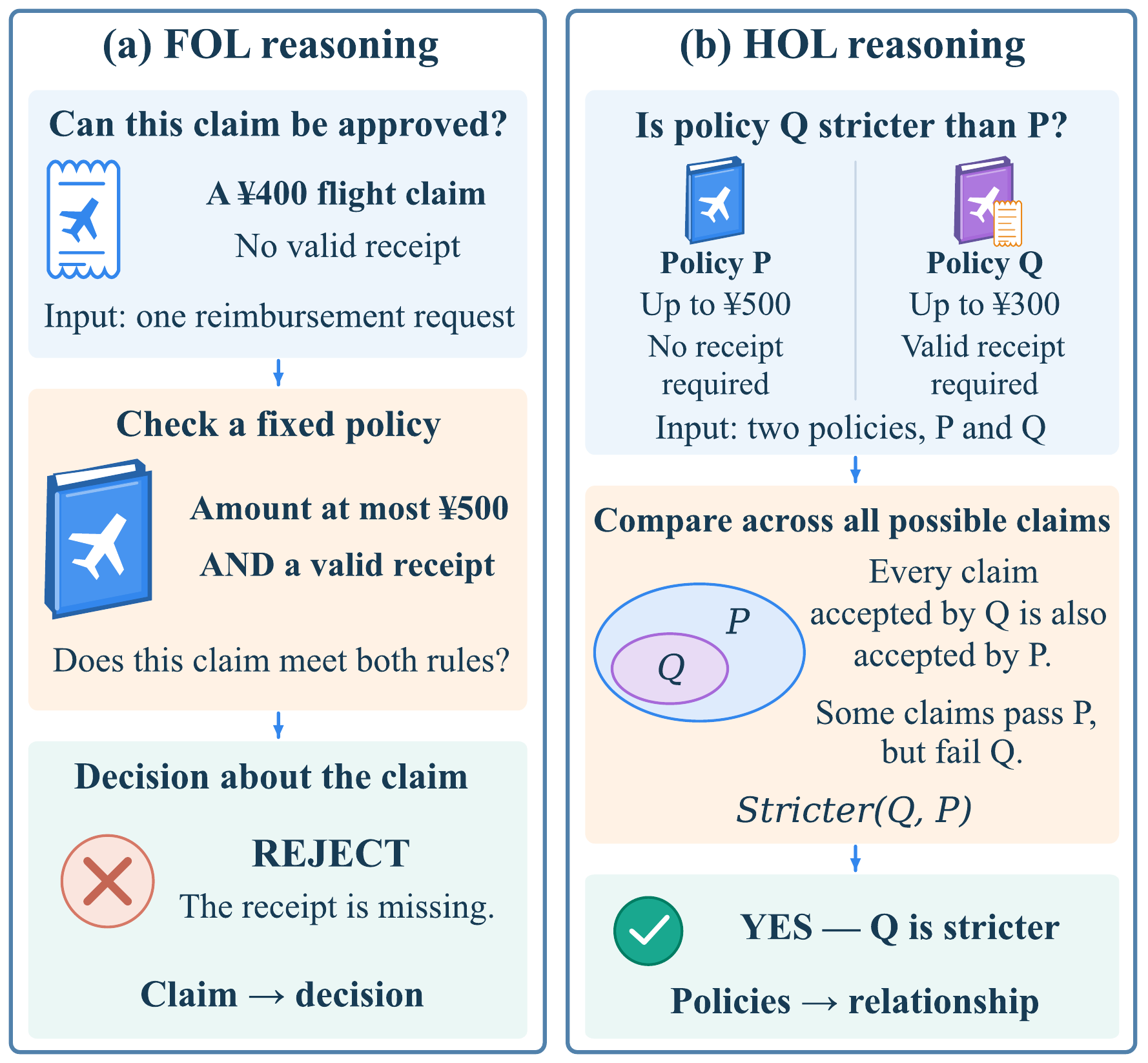}
    \caption{
    Illustration of the difference between object-level deduction in first-order logic (FOL) and higher-order logic (HOL) reasoning. FOL reasoning evaluates individual claims under fixed policies, whereas HOL reasoning can treat policies as predicates and reason about their relationships.
    }
    \vspace{-1em}
    \label{fig:fol-hol-comparison}
\end{figure}

Recent benchmarks have extensively evaluated the logical reasoning abilities of LLMs. Most existing datasets focus on object-level deduction, where models reason over individual entities and their relations under fixed predicates and rules~\citep{folio,proofwriter,proverqa}. This setting does not capture many real-world problems in which the rules or decision procedures themselves must be reasoned over, such as resolving competing legal rules or selecting, combining, and applying policies with exceptions in financial decision-making. As shown in Figure~\ref{fig:fol-hol-comparison}, higher-order reasoning therefore goes beyond applying fixed predicates to individual objects: models must reason over the rules and procedures that govern the decision itself. 
A second challenge concerns how such reasoning should be evaluated. Most existing logical reasoning benchmarks provide either no verified reasoning traces or only a single reference derivation, although the same conclusion may admit multiple valid reasoning paths. Evaluating model-generated reasoning against a single reference can therefore penalize alternative valid derivations or fail to distinguish them from genuine reasoning errors. Consequently, both higher-order reasoning and fine-grained evaluation of its intermediate reasoning processes remain insufficiently explored in existing LLM benchmarks.

\begin{figure*}
\includegraphics[width=\linewidth]{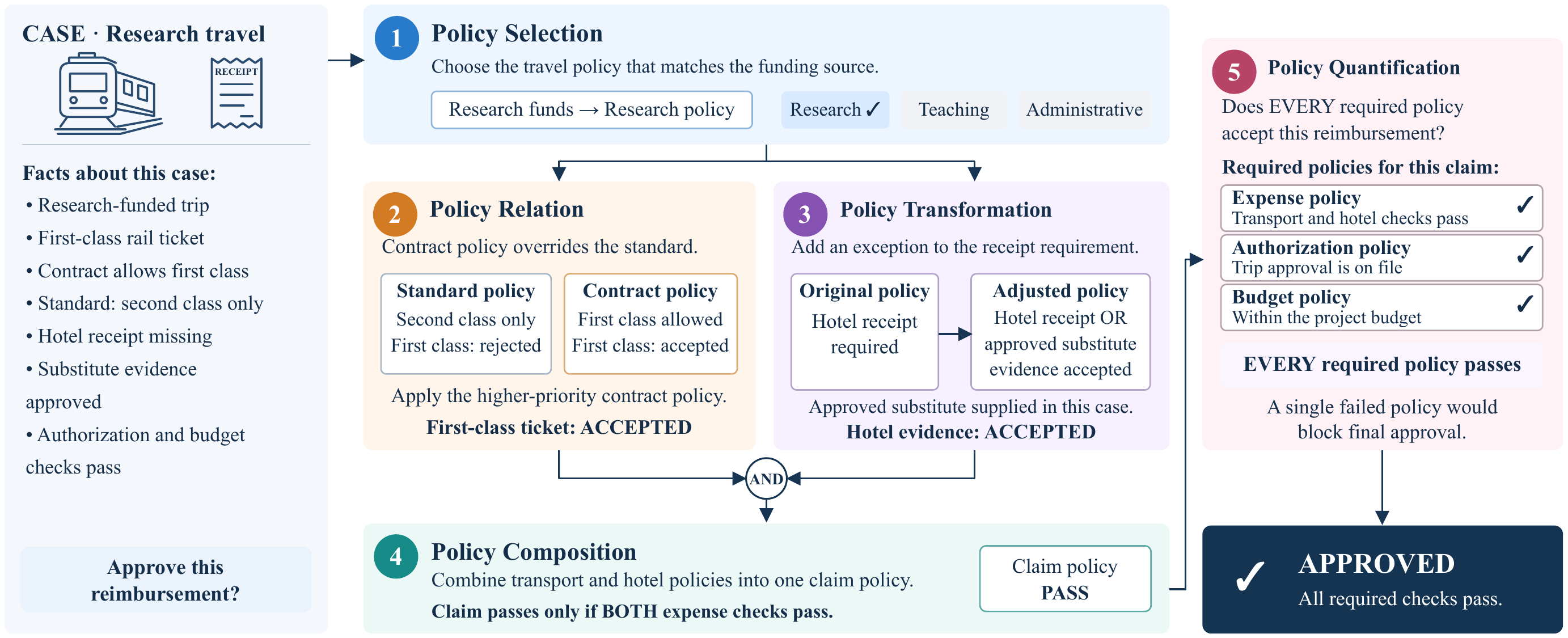}
\caption{Five policy-level reasoning operations illustrated through a case. Policy selection identifies the applicable research-travel policy. Policy relation gives the contract’s first-class travel allowance priority over the standard restriction, while policy transformation extends the receipt requirement to accept approved substitute evidence. Policy composition combines the transport and hotel policies into a single claim policy. Policy quantification checks whether every required policy, including expense, authorization, and budget policies, is satisfied.}
\label{fig:HOLogicBench_examples}
\end{figure*}

To address these gaps, we introduce \textbf{HOLogicBench}, a systematic benchmark for higher-order symbolic reasoning in LLMs. \textsc{HOLogicBench} organizes evaluation around recurring higher-order reasoning patterns in which rules, predicates, policies, and decision procedures can be related, selected, transformed, or composed. Each pattern is instantiated symbolically, verified in Isabelle/HOL, and then grounded in realistic rule-governed scenarios while preserving its intended logical structure. 
Beyond proof-verified answers and HOL formalizations, we develop an algorithm that enumerates all minimal sufficient derivations under the benchmark semantics, with a formal correctness guarantee, enabling reliable evaluation and fine-grained diagnosis of model-generated reasoning. Figures~\ref{fig:HOLogicBench_examples} and~\ref{fig:HOLogicBench_examples_text} illustrate representative examples.

We evaluate 16 configurations of open-source and proprietary LLMs on \textsc{HOLogicBench}, with an average accuracy of 53.86\% and a best performance of 65.99\%. 
Our analyses reveal substantial variation across higher-order reasoning types, limited alignment of LLM-generated reasoning with valid derivations, and vulnerability to minimal counterfactual perturbations. Further error analysis identifies distinct failure patterns across reasoning types. Together, these findings highlight persistent challenges in reliably applying higher-order rules and generating reasoning processes that adequately justify model conclusions.
Our contributions are summarized as follows:
\begin{itemize}
    \item We introduce \textsc{HOLogicBench}, a systematic benchmark for higher-order rule and policy reasoning in LLMs.
    
    \item We develop a sound algorithm for enumerating minimal sufficient derivations under the benchmark's inference semantics, enabling fine-grained evaluation of reasoning validity.
    
    \item We evaluate 16 configurations of open-source and proprietary LLMs on \textsc{HOLogicBench}, revealing substantial room for improvement, limited alignment with valid derivations, vulnerability to minimal counterfactual perturbations, and distinct failure patterns across higher-order reasoning types.
\end{itemize}

\begin{table}[t]
\centering
\small
\setlength{\tabcolsep}{3pt}
\renewcommand{\arraystretch}{1.15}
\begin{tabularx}{\columnwidth}{@{}>{\raggedright\arraybackslash}Xcccc@{}}
\toprule
\textbf{Benchmark}
& \textbf{\makecell{HOL\\types}}
& \textbf{\makecell{Solver-backed\\answers}}
& \textbf{\makecell{Proof\\traces}}
& \textbf{\makecell{Multiple\\proofs}} \\
\midrule
ProofWriter & \textcolor{red!80!black}{\xmark} & \textcolor{green!50!black}{\cmark} & \textcolor{green!50!black}{\cmark} & \textcolor{green!50!black}{\cmark} \\
FOLIO & \textcolor{red!80!black}{\xmark} & \textcolor{green!50!black}{\cmark} & \textcolor{red!80!black}{\xmark} & \textcolor{red!80!black}{\xmark} \\
SARA & \textcolor{red!80!black}{\xmark} & \textcolor{green!50!black}{\cmark} & \textcolor{red!80!black}{\xmark} & \textcolor{red!80!black}{\xmark} \\
BoardgameQA & \textcolor{red!80!black}{\xmark} & \textcolor{red!80!black}{\xmark} & \textcolor{green!50!black}{\cmark} & \textcolor{red!80!black}{\xmark} \\
RuleArena & \textcolor{red!80!black}{\xmark} & \textcolor{red!80!black}{\xmark} & \textcolor{red!80!black}{\xmark} & \textcolor{red!80!black}{\xmark} \\
\midrule
\textbf{HOLogicBench} & \textcolor{green!50!black}{\cmark} & \textcolor{green!50!black}{\cmark} & \textcolor{green!50!black}{\cmark} & \textcolor{green!50!black}{\cmark} \\
\bottomrule
\end{tabularx}
\caption{Comparison with representative benchmarks. \textcolor{green!50!black}{\cmark}: provided; \textcolor{red!80!black}{\xmark}: not reported under these criteria. HOL types denotes an explicit policy-level taxonomy; solver-backed answers use formal logic engines. Proof traces are reference logical derivations; multiple proofs denotes alternative derivations for the same answer.}
\label{tab:benchmark-positioning}
\end{table}

\begin{table*}[t]
\centering
\small
\setlength{\tabcolsep}{4pt}
\renewcommand{\arraystretch}{1.12}
\begin{tabularx}{\textwidth}{p{1.0cm} p{5.2cm} p{3.0cm} X}
\toprule
\textbf{HOL Type} &
\textbf{Typed HOL Schema} &
\textbf{Higher-Order Role} &
\textbf{Real-World Example} \\
\midrule

\textbf{PR}
&
$\mathit{Rel}:
\mathsf{Policy}\rightarrow
\mathsf{Policy}\rightarrow
\mathsf{bool}$

e.g., $\mathit{Override}(P,Q)$,
$\mathit{Subsumes}(P,Q)$
&
Policies as relation arguments
&
An emergency order overrides an otherwise applicable municipal regulation.
\\

\textbf{PT}
&
$\mathit{Transform}:
\mathsf{Policy}\rightarrow
\mathsf{Policy}$

$P'=\mathit{Transform}(P),\; P'(x)?$
&
Policy as input and output
&
A medical exception modifies the travel policy used to decide whether business-class travel is reimbursable.
\\

\textbf{PC}
&
$\mathit{Compose}:
\mathsf{Policy}\rightarrow
\mathsf{Policy}\rightarrow
\mathsf{Policy}$

$P'=\mathit{Compose}(P,Q),\; P'(x)?$
&
Policies as inputs and output
&
A grant-funded claim must satisfy both the university policy and the sponsor's funding policy.
\\

\textbf{PS}
&
$\mathit{Select}:
\mathsf{Context}\rightarrow
\mathsf{Policy}$

$P=\mathit{Select}(c),\; P(x)?$
&
Policy as returned object
&
An employee's role and funding source determine which reimbursement policy applies.
\\

\textbf{PQ}
&
$\Phi:
\mathsf{Policy}\rightarrow
\mathsf{bool}$

$\forall P\in\Pi.\,\Phi(P)?$
&
Policy as quantified variable
&
A hospital audit checks whether every departmental access policy satisfies a mandatory privacy requirement.
\\

\bottomrule
\end{tabularx}

\caption{
HOL reasoning types considered in \textsc{HOLogicBench}.
$\mathsf{Policy}\triangleq\mathcal{X}\rightarrow\mathsf{bool}$.
PR: Policy Relation; PT: Policy Transformation; PC: Policy Composition;
PS: Policy Selection; PQ: Policy Quantification.
}
\label{tab:hol-taxonomy}
\end{table*}

\section{Related Work}

Prior work has evaluated logical reasoning through benchmarks such as \emph{RuleTaker}~\citep{ruletaker}, \emph{ProofWriter}~\citep{proofwriter}, \emph{PrOntoQA}~\citep{prontoqa}, \emph{FOLIO}~\citep{folio}, and \emph{ProverQA}~\citep{proverqa}, while \emph{LogicBench}~\citep{parmar-etal-2024-logicbench}, \emph{Multi-LogiEval}~\citep{multi-logieval}, and \emph{LogiConBench}~\citep{chen2026logiconbench} cover broader reasoning patterns and multi-step compositions. However, these benchmarks primarily focus on deduction over entities under predefined rules, with limited coverage of higher-order operations over rules and policies. Moreover, most provide no or limited reference derivations. Although \emph{ProofWriter} enumerates multiple valid proofs, its bounded reasoning depth (<=5) setting makes exhaustive enumeration feasible. 
In contrast, \textsc{HOLogicBench} targets more complex higher-order rule dependencies and develops a sound and complete algorithm for enumerating minimal sufficient derivations, excluding redundant steps while preserving alternative valid reasoning paths.

A complementary line of work explores realistic rule-governed settings, including \emph{SARA}~\citep{sara}, \emph{LegalBench}~\citep{legalbench}, \emph{BoardgameQA}~\citep{boardgameqa}, \emph{RuleArena}~\citep{rulearena}, and \emph{FinRule-Bench}~\citep{finrulebench}. These benchmarks focus on domain-specific reasoning tasks rather than systematically evaluating higher-order rule operations. 
Instead, \textsc{HOLogicBench} combines five HOL reasoning types, formally verified answers, and fine-grained reasoning evaluation.

\section{HOLogicBench}

\begin{figure}[t]
    \centering
    \includegraphics[width=\linewidth]{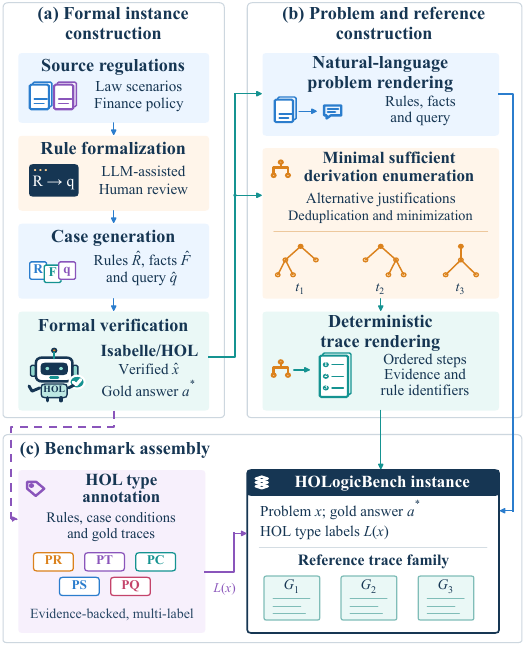}
    \caption{Overview of \textsc{HOLogicBench} construction. (a) Formal instances are generated and verified in Isabelle/HOL. (b) Verified instances are rendered as natural-language problems, while minimal sufficient derivations are enumerated and converted into reference traces. (c) Problems, gold answers, reference traces, and HOL type labels are assembled into the benchmark.}
    \vspace{-1em}
    \label{fig:construction}
\end{figure}

To evaluate both answer correctness and the validity of LLM-generated reasoning in higher-order tasks, we construct \textsc{HOLogicBench} through two connected stages, as illustrated in Figure~\ref{fig:construction}. First, \textbf{Formal Instance Construction} (\S~\ref{sec:dataset_construction}) formalizes real-world rules in Isabelle/HOL, generates cases involving higher-order reasoning operations, and verifies their gold answers. 
Each case is then presented as a natural-language problem with corresponding HOL type annotations. Building on these verified instances, \textbf{Reference Derivation and Trace Construction} (\S~\ref{sec:reference-construction}) enumerates minimal sufficient derivations and renders them into reference traces, supporting fine-grained evaluation of the LLM-generated traces. Finally, \textbf{Dataset Statistics} (\S~\ref{sec:dataset-statistics}) summarizes the benchmark's coverage and complexity.

\subsection{Formal Instance Construction}
\label{sec:dataset_construction}

We instantiate five HOL reasoning types within two rule-governed domains: law and finance. We represent a policy as a predicate $P:\mathcal{X}\rightarrow\mathsf{bool}$, where $\mathcal{X}$ denotes the domain of policy inputs. Higher-order operations take policies as arguments, return policies, or quantify over policies, as summarized in Table~\ref{tab:hol-taxonomy}.

Each natural-language instance is formulated as $x=(\mathcal{R},\mathcal{F},q)$, where $\mathcal{R}$ contains the governing rules and higher-order operators, $\mathcal{F}$ contains instance-specific facts, and $q$ is the query. Its Isabelle/HOL formalization $\hat{x}=(\hat{\mathcal{R}},\hat{\mathcal{F}},\hat{q})$ encodes the corresponding rules, facts, and query. The construction consists of three stages: (1) \textbf{Rule Formalization}, converting source regulations into typed Isabelle/HOL rule bases; (2) \textbf{Case Generation}, instantiating case-specific facts and queries involving target HOL operations; and (3) \textbf{Formal Verification}, verifying gold answers and logical consistency using Isabelle/HOL. The resulting formal instances are also rendered into natural-language problems for LLM evaluation.

\paragraph{Rule Formalization.}
We first select two real-world rule-governed scenarios, law and finance, to instantiate the higher-order reasoning types. For law, we construct 10 legal-style rule bundles from simplified Chinese Criminal Law text\footnote{\url{https://github.com/LawRefBook/Laws/blob/master/\%E5\%88\%91\%E6\%B3\%95/\%E5\%88\%91\%E6\%B3\%95.md}} and fictional regulatory systems, organized into atomic facts, intermediate concepts, legal provisions, and priority or exception relations. For finance, we formalize a 20-clause reimbursement policy from Fudan University\footnote{\url{http://www.it.fudan.edu.cn/Assets/userfiles/sys_eb538c1c-65ff-4e82-8e6a-a1ef01127fed/files/ggfw/\%E8\%B4\%A2\%E5\%8A\%A1\%E6\%8A\%A5\%E9\%94\%80\%E4\%BA\%8C\%E5\%8D\%81\%E6\%9D\%A1.pdf}} into typed rules over personnel attributes and case conditions. The initial formalization $\hat{\mathcal{R}}$ is produced by GPT-5.3 and then manually verified and refined to ensure semantic fidelity and Isabelle/HOL compatibility. Detailed construction procedures are provided in Appendix~\ref{app:law_construction} and \ref{app:finance_construction}.

\paragraph{Case Generation.}
Given the formalized rule base $\hat{\mathcal{R}}$, we construct instance-specific facts $\hat{\mathcal{F}}$ and queries $\hat{q}$ to trigger one or more target HOL reasoning types in Table~\ref{tab:hol-taxonomy}. For law, we first select a target configuration of provisions, including the relevant priority or exception relations, and then back-solve the intermediate concepts and atomic facts required to activate the intended rule interactions. For finance, we instantiate the reimbursement rules with carefully selected personnel and case attributes to trigger the target policy operations, including scope-dependent policy selection, policy transformations, and compositions across multiple cases. Detailed generation procedures are provided in Appendix~\ref{app:law_construction} and \ref{app:finance_construction}.

% \textcolor{red}{polished until here}

\paragraph{Formal Verification.}
For each generated formal instance $\hat{x}$, we use Isabelle/HOL to verify that the encoded rules and facts yield the target decision or value specified by $\hat{q}$, obtaining a proof-verified ground-truth answer $a^\star$. We additionally check theory compilation, consistency among rules, facts, and queries, and the absence of contradictory case facts. Verification details are provided in Appendix~\ref{app:law_construction} and ~\ref{app:finance_construction}.

% \paragraph{HOL Type Annotation.}
% {\color{red}
% We annotate each case $x$ with a label set $L(x)\subseteq\{\mathrm{PR},\mathrm{PT},\mathrm{PC},\mathrm{PS},\mathrm{PQ}\}$ based on the policy operations involved in the task. Reproducible annotation scripts inspect the rule definitions, case conditions, and available gold traces to identify case-relevant policy relations, transformations, compositions, selections, and quantification. Each assigned label is accompanied by evidence linking it to source definitions and conditions. Labels describe policy-level reasoning semantics rather than surface features such as arbitrary factual conjunctions, and multiple labels are allowed when operations co-occur. Cases without a qualifying operation retain an empty label set. This procedure supports evaluation by reasoning type; Appendix~\ref{app:hol-type-annotation} provides the detailed criteria and validation checks.
% }

\begin{figure*}[t]
\centering
\begin{minipage}[t]{0.715\textwidth}
\vspace*{0pt}
    \centering
\small
\setlength{\tabcolsep}{2.5pt}
\resizebox{\textwidth}{!}
{%
\begin{tabular}{lrrrrrr}
\toprule
\textbf{Statistic} & \textbf{PR} & \textbf{PT} & \textbf{PC} & \textbf{PS} & \textbf{PQ} & \textbf{Overall} \\
\midrule
Number of cases & 268 & 265 & 1{,}015 & 998 & 300 & 1{,}379 \\
Case length (tokens, min--max) & 41--717 & 45--1{,}001 & 169--1{,}008 & 169--1{,}008 & 32--126 & 32--1{,}008 \\
Case length (tokens, mean) & 87.65 & 171.54 & 381.92 & 381.57 & 75.54 & 308.54 \\
\midrule
Rules used per case (min--max) & 3--35 & 3--83 & 2--83 & 4--83 & 3--19 & 1--83 \\
Rules used per case (mean) & 7.39 & 11.97 & 25.11 & 25.74 & 7.91 & 20.88 \\
Reasoning paths per case (mean) & 4.95 & 787.61 & 258.30 & 262.65 & 8.92 & 192.11 \\
Reasoning depth (mean) & 6.32 & 6.71 & 6.70 & 6.89 & 6.29 & 6.55 \\
Reasoning steps (mean) & 12.00 & 19.20 & 39.96 & 40.86 & 12.83 & 32.98 \\
\midrule
GT trace length (tokens, mean) & 277.77 & 3{,}285.38 & 9{,}052.89 & 9{,}251.75 & 165.85 & 6{,}836.90 \\
LLM output length (tokens, mean) & 518.07 & 505.26 & 442.13 & 444.79 & 505.41 & 445.35 \\
\bottomrule
\end{tabular}}
\end{minipage}
\hfill
\begin{minipage}[t]{0.275\textwidth}
\vspace*{0pt}
    \centering
    \includegraphics[width=\linewidth]{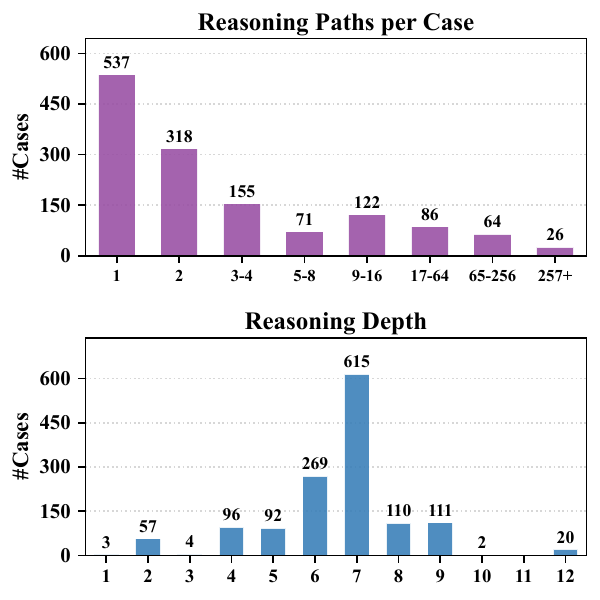}
    \label{fig:paths_depth_distribution}
\end{minipage}
\vspace{-2em}
\caption{Dataset statistics of \textsc{HOLogicBench}. Left: Case counts, input lengths, and reasoning characteristics across five overlapping HOL categories and the overall benchmark. Right: Distributions of reasoning paths per case (top) and reasoning depth (bottom) across all 1,379 cases.}
\label{tab:dataset_statistics}
\vspace{-1em}
\end{figure*}

\subsection{Reference Derivation and Trace Construction}
\label{sec:reference-construction}
\label{sec:reference-enumeration}
Building on the formally verified instances and their gold answers, we construct valid reference derivations that establish the corresponding conclusions.
Evaluating the reasoning process requires reference derivations that capture how a correct answer can be logically established. However, a single reference trace may fail to capture alternative valid reasoning paths, potentially penalizing correct derivations that differ from the reference. Conversely, enumerating all possible derivations can introduce redundant or arbitrarily long reasoning paths with unnecessary intermediate steps. To address these challenges, we construct a set of \emph{minimal sufficient derivations} for each verified answer $a^\star$, capturing alternative valid reasoning paths while excluding unnecessary steps.

\paragraph{Minimal Sufficient Derivation Enumeration.}
Let $t$ denote a symbolic derivation tree explaining $a^\star$, and let $\mathcal{D}_{\min}(x)$ be the set of all minimal sufficient trees under the criterion in Appendix~\ref{app:tree-enumeration}. Starting from $a^\star$, we recursively expand the formalized rules to enumerate all permitted justifications, combining jointly required premises while preserving alternative sufficient explanations. Structural deduplication and global minimization yield the algorithm's output $\operatorname{EnumMin}(\hat{x})$.
The resulting symbolic derivation trees are then flattened and rendered into reference reasoning traces, as described in Appendix~\ref{app:trace-rendering}. The following theorem establishes the completeness of the enumeration algorithm under the specified inference semantics:

\begin{theorem}
\label{thm:tree-enumeration}
Assuming a finite, acyclic explanation specification that faithfully encodes the permitted rule applications and branch priorities, the enumeration algorithm returns exactly all minimal sufficient derivations:
\[
    \operatorname{EnumMin}(\hat{x})=\mathcal{D}_{\min}(x).
\]
\end{theorem}

The proof and the complete enumeration algorithm are provided in Appendix~\ref{app:tree-enumeration}.

\subsection{Dataset Statistics}
\label{sec:dataset-statistics}

Figure~\ref{tab:dataset_statistics} summarizes the statistics of \textsc{HOLogicBench}. The benchmark contains 1{,}379 instances covering all five higher-order reasoning types in Table~\ref{tab:hol-taxonomy}: 268 involve Policy Relation (PR), 265 Policy Transformation (PT), 1{,}015 Policy Composition (PC), 998 Policy Selection (PS), and 300 Policy Quantification (PQ). These categories are non-exclusive, as a single instance may involve multiple higher-order reasoning types. Reasoning depth\footnote{Reasoning depth is the maximum number of named derived-rule
applications along a fact-to-target chain in the exported explanation
system; Boolean and arithmetic operations do not add a rule layer.} ranges from 1 to 12, providing substantial variation in reasoning complexity.

\section{Experiments}

% \subsection{Settings}

\paragraph{Models.}
We evaluate 16 model configurations on \textsc{HOLogicBench}, as reported in Table~\ref{tab:main_results}.
The evaluated models include DeepSeek-V3.2~\citep{deepseekai2025deepseekv32}, DeepSeek-R1~\citep{guo2025deepseekr1}, DeepSeek-V4-Flash, DeepSeek-V4-Pro, Qwen3-30B-Instruct and Qwen3-30B-Thinking~\citep{yang2025qwen3}, Qwen3.6-Flash, MiniMax-M2.5~\citep{minimax2026m25}, MiniMax-M2.7, MiniMax-M3~\citep{minimax2026m3}, GLM-5.3-Flash~\citep{zai2026glm53flash}, Kimi-K2.6~\citep{moonshot2026kimik26}, GPT-5.4-Mini~\citep{openai2026gpt54mini}, and Gemini-3.1-Flash~\citep{google2026gemini31flashlite}.
For DeepSeek-V4-Flash and MiniMax-M3, we evaluate both thinking (T) and non-thinking (NT) modes following their official configurations; for Qwen3-30B, we include the Instruct (I) and Thinking (T) variants.

\begin{table*}[t]
\centering
\small
\setlength{\tabcolsep}{2.5pt}
\renewcommand{\arraystretch}{1.14}
\begin{adjustbox}{max width=\linewidth}
\begin{tabular}{@{}l|rrr|rrr|rrr|rrr|rrr|rrr@{}}
\toprule
\textbf{Model} & \multicolumn{3}{c|}{\textbf{PR}} & \multicolumn{3}{c|}{\textbf{PT}} & \multicolumn{3}{c|}{\textbf{PC}} & \multicolumn{3}{c|}{\textbf{PS}} & \multicolumn{3}{c|}{\textbf{PQ}} & \multicolumn{3}{c}{\textbf{Overall}} \\
% \cmidrule(lr){2-4}\cmidrule(lr){5-7}\cmidrule(lr){8-10}\cmidrule(lr){11-13}\cmidrule(lr){14-16}\cmidrule(lr){17-19}
 & Acc. & B-F1 & SA-P & Acc. & B-F1 & SA-P & Acc. & B-F1 & SA-P & Acc. & B-F1 & SA-P & Acc. & B-F1 & SA-P & Acc. & B-F1 & SA-P \\
\midrule
DeepSeek-V3.2 & 83.21 & 0.7721 & 0.6195 & 69.43 & 0.7602 & 0.5980 & 37.24 & 0.7306 & 0.5976 & 37.07 & 0.7300 & 0.5987 & 85.00 & 0.7749 & 0.6299 & 49.67 & 0.7400 & 0.6070 \\
DeepSeek-R1 & 74.25 & 0.7628 & 0.6206 & 69.43 & 0.7540 & 0.5991 & 45.02 & 0.7352 & 0.5995 & 43.49 & 0.7339 & 0.5981 & 77.00 & 0.7654 & 0.6318 & 53.08 & 0.7414 & 0.6083 \\
DeepSeek-V4-Pro & 71.64 & 0.7802 & 0.6072 & 60.00 & 0.7628 & 0.5857 & 34.88 & 0.7246 & 0.5851 & 33.27 & 0.7234 & 0.5849 & 74.33 & 0.7833 & 0.6185 & 45.11 & 0.7372 & 0.5947 \\
Qwen3.6-Flash & \textbf{87.31} & 0.7778 & 0.6141 & \textbf{77.36} & 0.7615 & 0.5991 & 50.54 & 0.7249 & 0.6054 & 49.90 & 0.7240 & 0.6057 & \textbf{88.33} & 0.7805 & 0.6253 & 59.97 & 0.7367 & 0.6118 \\
MiniMax-M2.5 & 79.10 & 0.7658 & 0.6145 & 70.19 & 0.7542 & 0.5925 & 47.29 & 0.7324 & 0.6036 & 46.89 & 0.7311 & 0.6027 & 81.00 & 0.7688 & 0.6260 & 56.20 & 0.7400 & 0.6102 \\
MiniMax-M2.7 & 74.25 & 0.7627 & 0.6145 & 65.66 & 0.7528 & 0.5924 & 39.31 & 0.7364 & 0.6013 & 37.88 & 0.7359 & 0.6018 & 76.00 & 0.7650 & 0.6273 & 48.73 & 0.7424 & 0.6091 \\
GLM-5.3-Flash & 77.99 & 0.7762 & 0.6025 & 72.83 & 0.7623 & 0.5796 & \textbf{61.28} & 0.7350 & 0.5864 & \textbf{59.92} & 0.7342 & 0.5869 & 80.00 & 0.7797 & 0.6147 & \textbf{65.99} & 0.7446 & 0.5947 \\
Kimi-K2.6 & 78.36 & 0.7775 & \textbf{0.6307} & 68.30 & 0.7642 & 0.6095 & 55.17 & 0.7335 & 0.6054 & 53.91 & 0.7328 & 0.6058 & 81.00 & 0.7806 & \textbf{0.6413} & 61.86 & 0.7434 & 0.6154 \\
GPT-5.4-Mini & 79.85 & \textbf{0.7907} & 0.6206 & 67.92 & 0.7709 & \textbf{0.6095} & 34.88 & 0.7156 & \textbf{0.6139} & 34.17 & 0.7150 & \textbf{0.6157} & 81.33 & \textbf{0.7931} & 0.6278 & 46.19 & 0.7325 & \textbf{0.6188} \\
Gemini-3.1-Flash & 84.70 & 0.7891 & 0.6095 & 72.83 & \textbf{0.7723} & 0.5952 & 43.94 & 0.7205 & 0.5874 & 43.79 & 0.7193 & 0.5877 & 85.67 & 0.7906 & 0.6188 & 54.68 & 0.7354 & 0.5962 \\
\midrule
DeepSeek-V4-Flash-NT & 77.61 & 0.7788 & 0.6051 & 68.30 & 0.7635 & 0.5871 & 41.48 & 0.7295 & 0.5943 & 40.48 & 0.7287 & 0.5938 & 79.67 & 0.7816 & 0.6168 & 51.27 & 0.7405 & 0.6012 \\
DeepSeek-V4-Flash-T & 70.52 & 0.7746 & 0.6053 & 58.87 & 0.7601 & 0.5845 & 44.43 & 0.7326 & 0.5977 & 42.69 & 0.7318 & 0.5987 & 73.00 & 0.7779 & 0.6177 & 51.92 & 0.7423 & 0.6041 \\
\midrule
Qwen3-30B-I & 83.58 & 0.7850 & 0.6212 & 71.70 & 0.7678 & 0.5985 & 36.16 & 0.7255 & 0.5877 & 35.87 & 0.7253 & 0.5883 & 85.00 & 0.7882 & 0.6317 & 48.51 & 0.7389 & 0.5993 \\
Qwen3-30B-T & 82.09 & 0.7730 & 0.6281 & 71.32 & 0.7609 & 0.6050 & 47.00 & 0.7266 & 0.5923 & 46.09 & 0.7253 & 0.5928 & 84.00 & 0.7752 & 0.6402 & 56.42 & 0.7369 & 0.6056 \\
\midrule
MiniMax-M3-NT & 83.58 & 0.7796 & 0.6051 & 70.57 & 0.7653 & 0.5834 & 39.11 & 0.7372 & 0.5932 & 37.68 & 0.7365 & 0.5924 & 85.00 & 0.7830 & 0.6172 & 50.25 & 0.7470 & 0.6005 \\
MiniMax-M3-T & 82.09 & 0.7754 & 0.5990 & 74.34 & 0.7625 & 0.5773 & 54.78 & \textbf{0.7397} & 0.5933 & 53.51 & \textbf{0.7390} & 0.5931 & 83.67 & 0.7791 & 0.6112 & 61.86 & \textbf{0.7481} & 0.5995 \\
\midrule
\rowcolor{blue!8}
\textbf{Average} & 79.38 & 0.7763 & 0.6136 & 69.32 & 0.7622 & 0.5935 & 44.53 & 0.7300 & 0.5965 & 43.54 & 0.7291 & 0.5967 & 81.25 & 0.7792 & 0.6248 & 53.86 & 0.7405 & 0.6048 \\
\bottomrule
\end{tabular}
\end{adjustbox}
\caption{Results on \textsc{HOLogicBench} by HOL reasoning type. Each category reports accuracy (Acc., \%), BERTScore-F1 (B-F1), and ROSCOE semantic-alignment precision (SA-P). I, T, and NT denote Instruct, Thinking, and Non-thinking. Bold marks the best model in each column (excluding Average).} 
% Categories overlap; Overall averages all 1,379 cases once, including 20 outside the five types. Average is the unweighted mean across models.}
\vspace{-1em}
\label{tab:main_results}
\end{table*}

\subsection{Evaluation Protocol}
\label{sec:reasoning-trace-evaluation}

We evaluate both final-answer correctness and the alignment of LLM-generated reasoning with valid reference derivations. For each case $x_i$, the model produces an answer $a_i$ and reasoning trace $\tau_i$, while \textsc{HOLogicBench} provides a verified answer $a_i^\star$ and reference family $\mathfrak{G}_{x_i}$.

\textbf{Answer Accuracy (Acc.).}
We report exact-match accuracy against Isabelle/HOL-verified answers, using task-specific answer normalization.

\textbf{BERTScore-F1 (B-F1).}
We use BERTScore-F1~\citep{bertscore} to measure semantic similarity between generated and reference reasoning texts through contextual token representations.

\textbf{ROSCOE-SA-P (SA-P).}
We adopt ROSCOE semantic-alignment precision~\citep{ROSCOE} to measure step-level similarity. Given generated steps $\tau=(s_1,\ldots,s_T)$, reference steps $\mathcal{G}=(g_1,\ldots,g_K)$, and normalized sentence embeddings $e(\cdot)$,
\begin{equation}
\operatorname{SA\text{-}P}(\tau,\mathcal{G})=
\frac{1}{T}\sum_{j=1}^{T}\max_{1\leq k\leq K}
e(s_j)^\top e(g_k).
\end{equation}

\textbf{Multi-reference Aggregation.}
To account for alternative valid derivations, we evaluate each reasoning metric against all reference traces and retain the highest score:
\begin{equation}
m_i^\star=\max_{\mathcal{G}\in\mathfrak{G}_{x_i}}
m(\tau_i,\mathcal{G}),\qquad
\overline{m}=\frac{1}{N}\sum_{i=1}^{N}m_i^\star,
\label{eq:multireference-score}
\end{equation}
where $m\in\{\operatorname{B\text{-}F1},\operatorname{SA\text{-}P}\}$. Each metric independently selects the best-matching complete reference path, accommodating alternative valid derivations without combining steps across paths. 
% Details are provided in Appendix~\ref{app:reasoning-implementation}.
Model configurations, reference aggregation, and similarity-metric
implementations are detailed in Appendix~\ref{app:reasoning-implementation}.
Implementation details for all five analysis experiments are provided
in Appendix~\ref{app:analysis-implementation}.

\subsection{Main Results}

Table~\ref{tab:main_results} presents the main results on \textsc{HOLogicBench}, with the following key findings.

\paragraph{Current LLMs exhibit substantial room for improvement in higher-order reasoning.}
Across all evaluated models, the average accuracy is only 53.86\%, with the best-performing model reaching 65.99\%. Even the strongest model therefore fails on over one-third of the cases, indicating that \textsc{HOLogicBench} remains challenging for current LLMs. These results suggest that, despite recent advances in LLM reasoning capabilities, reliably applying complex rules, policies, and decision procedures remains an open challenge.

\paragraph{Performance varies substantially across cases involving different HOL reasoning types.}
Models achieve relatively high average accuracy on cases involving PR (79.38\%), PT (69.32\%), and PQ (81.25\%), but perform considerably worse on cases involving PC (44.53\%) and PS (43.54\%), despite comparable reasoning depths across categories.
This gap suggests that cases involving policy composition and context-dependent policy selection pose greater challenges, potentially due to the need to coordinate multiple interacting rules and conditions. 
Interestingly, the best-performing models differ across reasoning types, with no model outperforming others across all categories. 
% This highlights the heterogeneous reasoning capabilities of current LLMs, which may not be fully reflected by overall accuracy alone.

\begin{figure*}[t]
    \centering
    % Left: transposed table
    \begin{minipage}[t]{0.32\textwidth}
        \centering
        \vspace{0pt}
        \small
        \setlength{\tabcolsep}{2.5pt}
        \renewcommand{\arraystretch}{1.12}
        \begin{adjustbox}{max width=\linewidth}
        \begin{tabular}{@{}lccc@{}}
            \toprule
            \textbf{Metric (\%)}
            & \shortstack{\textbf{MiniMax}\\\textbf{M3-T}}
            & \shortstack{\textbf{MiniMax}\\\textbf{M3-NT}}
            & \shortstack{\textbf{GLM-5.3}\\\textbf{Flash}} \\
            \midrule
            O-Acc $\uparrow$
            & \textbf{58.40} & 56.40 & 54.80 \\
            CF-Acc $\uparrow$
            & \textbf{62.00} & 54.80 & 58.00 \\
            CRS $\uparrow$
            & \textbf{86.99} & 75.89 & 86.86 \\
            Ctrl-Fail $\downarrow$
            & 23.60 & 34.00 & \textbf{22.80} \\
            All-3 $\uparrow$
            & \textbf{47.60} & 38.00 & 42.80 \\
            \bottomrule
        \end{tabular}
        \end{adjustbox}
        \captionof{table}{Counterfactual intervention results.
        O/CF-Acc: original/decisive-variant accuracy; CRS: CF-Acc on originally correct cases; Ctrl-Fail: prediction changes or missing responses under control interventions; All-3: all three versions correct.}
        % \captionof{table}{Counterfactual intervention results (\%).
        % CRS is conditioned on originally correct cases; other metrics
        % use all 250 cases. All-3 measures correctness across all three
        % variants. Missing responses count as failures.}
        \label{tab:counterfactual}
    \end{minipage}
    \hfill
    % Middle: depth versus accuracy
    \begin{minipage}[t]{0.325\textwidth}
        \centering
        \vspace{0pt}
        \includegraphics[width=\linewidth]{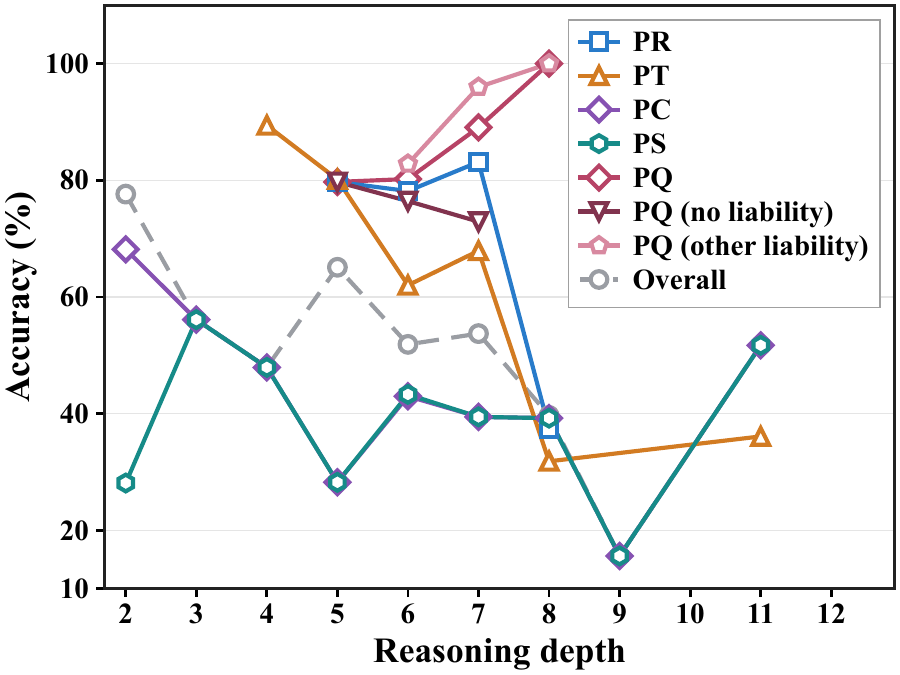}
        \caption{Accuracy versus reasoning depth across HOL categories,
        averaged over 16 models. PQ is further split by liability outcome.}
        \label{fig:depth_accuracy}
    \end{minipage}
    \hfill
    % Right: failure mechanisms
    \begin{minipage}[t]{0.325\textwidth}
        \centering
        \vspace{0pt}
        \includegraphics[width=\linewidth]{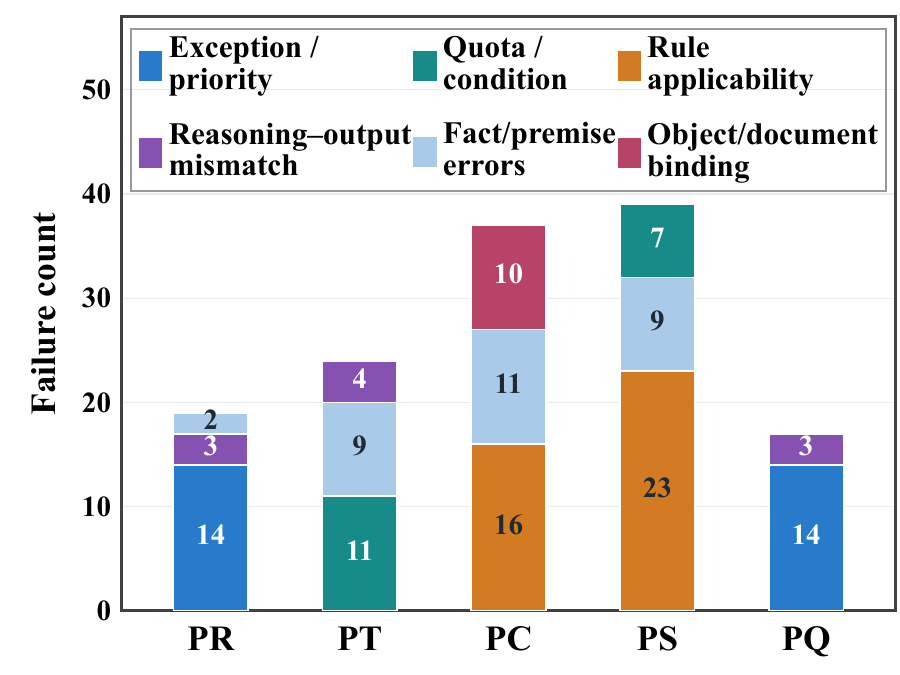}
        \caption{Common reasoning errors by HOL type, identified by
        GPT-6 Astra in 320 sampled outputs from 4 models.
        Labels show sample counts.}
        % \caption{Top failure mechanisms across HOL types.
        % Labels indicate sample counts; mechanisms may overlap.}
        \label{fig:hol_errors}
    \end{minipage}
\vspace{-1em}
\end{figure*}

\subsection{Reasoning Trace Analysis}
\label{sec:reasoning-analysis}

To further examine whether model-generated reasoning follows valid derivation paths, we evaluate five models on 500 sampled cases (100 per HOL reasoning type). To accommodate alternative valid derivations, we first select up to three of the most similar references from the enumerated minimal sufficient derivations. We then employ GPT-6-Luna to assess each candidate using four metrics: \emph{Derivation Precision} and \emph{Coverage}, measuring the proportions of generated and reference steps correctly aligned, respectively; \emph{Derivation F1}, their harmonic mean; and \emph{Critical-step Correctness}, measuring the proportion of critical derivation steps correctly executed. 
All metrics are reported against the reference with the highest Derivation F1. We validate the structural consistency of bidirectional step alignments
and retry invalid or incomplete judgments.

% We further employ a dual-blind judging procedure to filter unreliable assessments, with implementation details in Appendix~\ref{app:reasoning-implementation}.

As shown in Table~\ref{tab:reasoning}, \textbf{semantic similarity does not fully capture derivation-level alignment}. While BERTScore-F1 scores are nearly identical across models (75.54--76.06\%), derivation F1 reveals greater differences (22.64--27.82\%), revealing limited alignment with valid reasoning paths. 
More importantly, \textbf{correct answers are not always accompanied by complete and valid justifications of critical derivation steps}, with critical-step correctness ranging from only 17.72\% to 27.41\%. 
For example, MiniMax-M3-T and Gemini-3.1-Flash achieve comparable accuracy (69.00\% vs.\ 67.00\%) but differ substantially in critical-step correctness (27.41\% vs.\ 17.72\%). 
Interestingly, Gemini achieves the highest derivation precision but the lowest coverage, suggesting that its generated reasoning aligns with valid steps more selectively while leaving substantial derivation uncovered. 
These findings highlight the value of fine-grained evaluation in assessing LLM alignment with valid HOL reasoning traces.

\subsection{Counterfactual Intervention}
\label{sec:counterfactual}

To complement reasoning-trace alignment, we use counterfactual interventions to examine whether model decisions respond appropriately to changes in underlying HOL conditions. We conduct the interventions on 250 cases (50 per HOL reasoning type). 
% For each case, we construct two minimally modified variants, each changing only one rule or fact: a \emph{decisive intervention} that changes the original ground truth answer, and a \emph{control intervention} that leaves it unchanged.
For each case, we construct two variants that each change exactly one factual input field: a \emph{decisive intervention} that changes the verified answer, and a \emph{control intervention} that preserves it.

We report original-case accuracy (O-Acc) alongside three intervention metrics: (1) \emph{Counterfactual Accuracy} (CF-Acc), measuring accuracy on decisive variants; (2) \emph{Counterfactual Response Sensitivity} (CRS), measuring decisive-variant accuracy conditioned on the original answer being correct; and 
(3) \emph{Control Failure Rate} (Ctrl-Fail), measuring how often a control intervention changes a prediction that should remain unchanged. We also report All-3, the proportion of cases answered correctly in all three versions.

As shown in Table~\ref{tab:counterfactual}, \textbf{model decisions remain vulnerable to both decisive and non-decisive perturbations}.
Even among originally correct cases, models fail to correctly update their answers in 13.01\%--24.11\% of cases following a minimal intervention.
Moreover, despite the verified answer remaining unchanged, 22.80\%--34.00\% of control interventions lead models to change their predictions or fail to respond, revealing substantial decision instability.
Overall, consistent correctness across interventions remains limited, with only 38.00\%--47.60\% of cases answered correctly across all three versions.
These findings highlight the fragility of current LLMs in applying higher-order rules and conditions, with their decisions remaining sensitive to even minimal perturbations.

% \begin{table}[t]
% \centering
% \small
% % \setlength{\tabcolsep}{3pt}
% \setlength{\tabcolsep}{2.5pt}
% \renewcommand{\arraystretch}{1.12}
% \begin{adjustbox}{max width=\columnwidth}
% \begin{tabular}{@{}lccccc@{}}
% \toprule
% \textbf{Model} & \textbf{Acc.} & \textbf{Deriv-P} & \textbf{Coverage} & \textbf{Deriv-F1} & \textbf{Crit-C} \\
% \midrule
% MiniMax-M3-T     & \textbf{71.17} & 31.36 & \textbf{32.49} & \textbf{27.92} & \textbf{27.89} \\
% DeepSeek-V4-Pro  & 57.66 & 31.28 & 28.23 & 25.16 & 23.99 \\
% Qwen3-30B-I      & 66.22 & 28.12 & 28.36 & 24.63 & 24.08 \\
% GPT-5.4-Mini     & 61.71 & 30.34 & 24.88 & 23.61 & 21.26 \\
% Gemini-3.1-Flash & 70.50 & \textbf{35.73} & 20.89 & 23.55 & 18.44 \\
% \bottomrule
% \end{tabular}
% \end{adjustbox}
% \caption{Reasoning trace evaluation on 444 cases (\%). Acc.: answer accuracy; Deriv-P: derivation precision; Coverage: derivation recall; Deriv-F1: derivation F1; Crit-C: critical-step correctness.}
% \label{tab:reasoning}
% \end{table}

\begin{table}[t]
\centering
\small
\setlength{\tabcolsep}{2.5pt}
\renewcommand{\arraystretch}{1.12}
\begin{adjustbox}{max width=\columnwidth}
\begin{tabular}{@{}lccccc@{}}
\toprule
\textbf{Model} & \textbf{Acc.} & \textbf{Deriv-P} & \textbf{Coverage} & \textbf{Deriv-F1} & \textbf{Crit-C} \\
\midrule
MiniMax-M3-T     & \textbf{69.00} & 32.63 & \textbf{31.49} & \textbf{27.82} & \textbf{27.41} \\
DeepSeek-V4-Pro  & 55.00 & 32.11 & 27.07 & 24.72 & 23.33 \\
Qwen3-30B-I      & 62.40 & 28.70 & 26.92 & 23.89 & 23.14 \\
GPT-5.4-Mini     & 58.20 & 30.85 & 23.69 & 22.89 & 20.52 \\
Gemini-3.1-Flash & 67.00 & \textbf{35.89} & 19.87 & 22.64 & 17.72 \\
\bottomrule
\end{tabular}
\end{adjustbox}
\caption{Reasoning trace evaluation on 500 cases (\%). Acc.: answer accuracy; Deriv-P: derivation precision; Coverage: derivation recall; Deriv-F1: derivation F1; Crit-C: critical-step correctness.}
\label{tab:reasoning}
\vspace{-1em}
\end{table}

\subsection{Impact of Reasoning Complexity}

% To investigate how model performance varies with reasoning complexity, we analyze accuracy across different reference derivation lengths. As shown in Figure~\ref{fig:depth_accuracy}, \textbf{accuracy generally decreases as reference derivations become longer}, indicating that more complex reasoning cases pose greater challenges to current LLMs. 
% Interestingly, PQ exhibits an opposite trend, with accuracy increasing as derivation length grows. 
% Further examination reveals that the increasing accuracy in PQ is driven by the presence of applicable exemption rules, which allow models to reach the correct legal conclusion without resolving all competing rules. Figure~\ref{fig:depth_accuracy} shows that cases without applicable exemptions exhibit declining accuracy as derivation length increases, whereas cases with applicable exemptions show a different trend. These findings highlight that \textbf{reasoning difficulty depends not only on derivation length but also on the underlying rule structure and the availability of decisive rules}.

To investigate how model performance varies with reasoning complexity, we analyze accuracy across different reasoning depths. As shown in Figure~\ref{fig:depth_accuracy}, \textbf{accuracy generally decreases as reasoning depth increases}, indicating that cases requiring deeper reasoning pose greater challenges to current LLMs. 
Interestingly, PQ exhibits an opposite trend, with accuracy increasing as reasoning depth increases. 
% Further examination reveals that the increasing accuracy in PQ is driven by the presence of applicable exemption rules, which allow models to reach the correct legal conclusion without resolving all competing rules. 
% Figure~\ref{fig:depth_accuracy} shows that cases without applicable exemptions exhibit declining accuracy as reasoning depth increases, whereas cases with applicable exemptions show a different trend. 
To further examine this trend, we stratify cases by their verified liability outcomes. As shown in Figure~\ref{fig:depth_accuracy}, accuracy decreases with reasoning depth for no-liability cases but increases for cases with other outcomes. This contrasting pattern suggests that the relationship between reasoning depth and model performance in PQ depends on the liability outcome.
These findings highlight that models' behaviour depends not only on reasoning depth but also on the underlying rule structure and the availability of decisive rules.

\subsection{Impact of Thinking}
\label{sec:thinking-analysis}
% To examine whether the current thinking paradigm consistently improves reasoning across different HOL types, we compare three model pairs with and without thinking.
To examine the effects of thinking across HOL types, we compare three
paired model configurations using thinking and corresponding
non-thinking or instruct outputs.
As shown in Table~\ref{tab:thinking}, \textbf{thinking yields the largest improvements on cases involving policy composition (PC) and selection (PS), while showing limited or negative gains on other HOL reasoning types}. An analysis of 52 output pairs suggests that these improvements may stem from more accurate constraint application, numerical computation, and consistency between reasoning and final answers. However, thinking can also introduce unsupported assumptions or misapply exceptions. These findings highlight that thinking exhibits distinct effects across HOL reasoning types, reflecting differences in how models handle varying rule structures.

\begin{table}[t]
\centering
\small
\setlength{\tabcolsep}{4pt}
\renewcommand{\arraystretch}{1.12}
\begin{adjustbox}{max width=\columnwidth}
\begin{tabular}{@{}lrrrrrr@{}}
\toprule
\textbf{Model} & \textbf{PR} & \textbf{PT} & \textbf{PC} & \textbf{PS} & \textbf{PQ} & \textbf{Overall} \\
\midrule
DeepSeek-V4-Flash & -7.09 & -9.43 & +2.96 & +2.20 & -6.67 & +0.65 \\
Qwen3-30B         & -1.49 & -0.38 & +10.84 & +10.22 & -1.00 & +7.90 \\
MiniMax-M3        & -1.49 & +3.77 & +15.67 & +15.83 & -1.33 & +11.60 \\
\bottomrule
\end{tabular}
\end{adjustbox}
\caption{Accuracy differences between thinking and corresponding non-thinking or instruct configurations. Positive values indicate improvements with thinking.}
\label{tab:thinking}
\vspace{-1em}
\end{table}

\subsection{Error Analysis}
\label{sec:error-analysis}

To identify common failure modes in higher-order reasoning and assess the reliability of answer-level evaluation, we analyze 320 outputs from four representative models, sampling 12 incorrect and 4 correct predictions per model and HOL type. 
% We combine manual review with GPT-6 Astra to classify reasoning errors.
We use GPT-6 Astra annotations with a human review of a 64-output subset.
For each HOL type, we report the percentage of outputs exhibiting
each error mechanism among those with at least one confirmed error.

As shown in Figure~\ref{fig:hol_errors}, \textbf{different HOL reasoning types exhibit distinct failure patterns}. Errors in PR and PQ predominantly involve incorrect application of rule priorities and exceptions (73.7\% and 93.3\%), while PT failures frequently involve incorrect condition or numerical transformations (42.3\%). In contrast, PC and PS are particularly prone to incorrect policy applicability judgments and branch selection (44.4\% and 52.3\%).
Moreover, \textbf{correct answers do not necessarily indicate correct reasoning}. Among 80 originally correct predictions, 12 (15.0\%) still exhibit reasoning errors identified by the judge. 
These findings further highlight the importance of process-level evaluation in rigorous logical reasoning tasks, where correct conclusions must be supported by valid rule applications and derivations.
% More analysis can be found in the Appendix~\ref{}.

\section{Conclusion}

We introduce \textsc{HOLogicBench}, a formally verified benchmark covering five types of higher-order logical reasoning. Experiments reveal limitations in current LLMs, particularly on cases involving policy composition and selection. We enumerate minimal sufficient derivations to enable fine-grained evaluation of LLM-generated reasoning. Further analyses reveal limited derivation coverage, vulnerability to counterfactual perturbations, and distinct failure patterns across HOL reasoning types. These findings highlight the challenges of reliable higher-order reasoning and the importance of evaluating reasoning processes.

\newpage

\section*{Limitations}

Although \textsc{HOLogicBench} provides a testbed for higher-order symbolic reasoning, its current version focuses on two high-stakes domains: law and finance. 
These domains cover important higher-order reasoning patterns such as rule priority, exceptions, scope-conditioned decision procedures, and compositional rule application, but they do not exhaust the full range of higher-order reasoning required in real-world applications. 
Future work can extend \textsc{HOLogicBench} to additional domains such as medicine, scientific reasoning, program verification, and policy compliance, where rules, functions, constraints, and decision procedures also play central roles. 
We hope \textsc{HOLogicBench} can serve as a foundation for broader evaluations of faithful higher-order reasoning across more diverse tasks and domains.

\section*{Acknowledgement}

This work was supported in part by the Beijing Major Science and Technology Project under Contract No. Z251100008125054. This work was supported by the Beijing Academy of Artificial Intelligence (BAAI).

% \section*{Acknowledgments}

% This document has been adapted
% by Steven Bethard, Ryan Cotterell and Rui Yan
% from the instructions for earlier ACL and NAACL proceedings, including those for
% ACL 2019 by Douwe Kiela and Ivan Vuli\'{c},
% NAACL 2019 by Stephanie Lukin and Alla Roskovskaya,
% ACL 2018 by Shay Cohen, Kevin Gimpel, and Wei Lu,
% NAACL 2018 by Margaret Mitchell and Stephanie Lukin,
% Bib\TeX{} suggestions for (NA)ACL 2017/2018 from Jason Eisner,
% ACL 2017 by Dan Gildea and Min-Yen Kan,
% NAACL 2017 by Margaret Mitchell,
% ACL 2012 by Maggie Li and Michael White,
% ACL 2010 by Jing-Shin Chang and Philipp Koehn,
% ACL 2008 by Johanna D. Moore, Simone Teufel, James Allan, and Sadaoki Furui,
% ACL 2005 by Hwee Tou Ng and Kemal Oflazer,
% ACL 2002 by Eugene Charniak and Dekang Lin,
% and earlier ACL and EACL formats written by several people, including
% John Chen, Henry S. Thompson and Donald Walker.
% Additional elements were taken from the formatting instructions of the \emph{International Joint Conference on Artificial Intelligence} and the \emph{Conference on Computer Vision and Pattern Recognition}.

% Bibliography entries for the entire Anthology, followed by custom entries
%\bibliography{anthology,custom}
% Custom bibliography entries only

% \newpage

\bibliography{custom}

\appendix

\clearpage

The appendices follow the benchmark pipeline: Appendix~\ref{app:dataset-details}
describes dataset construction, type annotation, and examples;
Appendix~\ref{app:reference-details} presents the enumeration guarantee and
reference-trace rendering; Appendix~\ref{app:reasoning-implementation} documents
the main evaluation; Appendix~\ref{app:analysis-implementation} details the
analysis experiments; and Appendix~\ref{app:ethics-ai} covers ethics and AI use.

\section{Dataset Construction and Annotation}
\label{app:dataset-details}

\subsection{Law Dataset}
\label{app:law_construction}
The Law dataset evaluates reasoning through hierarchical rule application and
conflict resolution. Each rule bundle contains HOL definitions, natural-language
rules and cases, and solver-verified answers and reasoning traces. The following
pipeline expands the construction procedure in Section~\ref{sec:dataset_construction}.

\subsubsection{Construction Pipeline}

\paragraph{Step 1: Constructing rule bundles.}
We construct 10 self-contained bundles: five inspired by real law and five
fictional regulatory systems. The real-law-inspired bundles draw on selected
provisions and legal patterns from the Chinese Criminal Law text.
Rather than formalizing these materials verbatim, we rewrite and simplify them,
combining provisions and adding or removing conditions to support controlled
reasoning tasks. The fictional bundles follow a legal style but are independent
of real-world legal knowledge, reducing opportunities to answer through memorized
legal facts. Together, the final catalogs contain 285 numbered rules,
with 26--32 rules per bundle.

\paragraph{Step 2: Formalizing the rule hierarchy.}
Each bundle is represented in HOL using four layers: atomic case facts;
intermediate concepts defined over those facts; article-triggering conditions
and legal consequences; and conflict-resolution rules expressing priority and
exception-based defeat. The resulting dependency structure is
\emph{atomic facts $\rightarrow$ intermediate concepts $\rightarrow$ article
triggers $\rightarrow$ conflict resolution $\rightarrow$ final judgment}.
These layers support error localization: a model may misread a fact, fail to
establish an intermediate concept, trigger an inapplicable article, or resolve
a conflict incorrectly. The explicit dependencies distinguish these failures
from an incorrect final judgment alone.

\paragraph{Step 3: Generating and verifying cases.}
Structured metadata records predicate dependencies, article-trigger mappings,
and conflict relations. We use it to generate case plans specifying the
articles to be triggered, then back-solve sufficient intermediate concepts
and atomic facts. Each plan produces a symbolic case and a gold trace recording
triggered concepts and articles, priority and exception conflicts, and the
final applicable results. Isabelle/HOL verifies the corresponding derivations,
so gold answers follow from the formal rules rather than human answer annotation.
The case plan, symbolic specification, and verified trace together provide an
explicit record of how each intended conclusion is obtained.

\paragraph{Step 4: Rendering rules and cases.}
Once the symbolic rules and case plans are fixed, we render them in natural
language. Each bundle includes both legal-style provisions and interpretive
clauses linking atomic facts to intermediate concepts. Case descriptions express
only the selected atomic facts, without revealing intermediate concepts or
article-level conclusions. This preserves the correspondence with the symbolic
specification while leaving the model to infer the applicable rules and outcome.

\paragraph{Step 5: Defining evaluation questions.}
Each question asks for the final legal conclusion using only the supplied
rules and facts. Models provide a step-by-step trace with rule-number citations
and a final answer. The structured citations support automatic inspection of
rule use and reference-trace coverage alongside answer accuracy, allowing
correct conclusions and incomplete justifications to be assessed separately.

\subsubsection{Prompt Design and Quality Control}

\paragraph{Prompt design.}
A unified prompt specifies the permitted information sources, requests
step-by-step reasoning, requires each reasoning step to cite one rule number,
and reserves the final answer line for the legal conclusion. These constraints
allow free-form explanations while keeping traces and answers machine-parsable.

\paragraph{Difficulty control.}
\label{app:law_difficulty}
For each instance, we record
\begin{equation}
    d(x) = (\ell,\;|\mathcal{C}^\star|,\;|\mathcal{P}^\star|,\;|\mathcal{E}^\star|),
\end{equation}
where $\ell$ is the reasoning-chain length from $\mathcal{F}$ to the final
judgment, $\mathcal{C}^\star\subseteq\mathcal{C}$ contains the triggered
intermediate concepts, $\mathcal{P}^\star\subseteq\mathcal{P}$ the triggered
articles, and $\mathcal{E}^\star\subseteq\mathcal{E}$ the exceptions or
conflicts resolved along the chain. Controlling these dimensions during case-plan
selection in Step~3 supports difficulty stratification and coverage across
reasoning depths. The profile records both the length of the derivation and
the number of concepts, articles, and interactions involved.

\paragraph{Quality control.}
Symbolic checks verify the rule hierarchy, reject contradictory atomic facts,
and confirm that Isabelle theories compile and prove the intended conclusions.
We also check consistency among case plans, dataset records, and generated
Isabelle files. Manual review of the natural-language rules and cases removes
unnatural phrasing, obvious ambiguity, and accidental leakage of symbolic
predicate names. Finally, pilot outputs and recurring errors inform prompt
and parser refinements, particularly where correct final judgments must be
distinguished from incomplete or shortcut-based reasoning traces.

\subsection{Finance Dataset}
\label{app:finance_construction}
The Finance dataset is built from a formalized reimbursement policy, instantiated
HOL case templates, and solver-derived gold answers. Its five-stage pipeline
extends Section~\ref{sec:dataset_construction} with both individual and
compositional reimbursement tasks.

\subsubsection{Construction Pipeline}

\paragraph{Step 1: Formalizing reimbursement rules.}
We use Fudan University's 20-rule reimbursement policy. The first two rules
provide operational guidance; rules 3--11 and 13--20 specify reimbursement
conditions, including spending limits, approval requirements, and documentation.
GPT-5.3 assists in translating this business logic into HOL, followed by manual
inspection and correction to obtain an Isabelle rule base.

The rule base supports five task families:
\begin{enumerate}[label=(\roman*)]
  \item Reimbursement approval for individual expenses under rules 6--11 and
        13--19, covering office and printing, travel and overseas travel,
        conferences and training, family visits, labor, engineering and
        renovation, warranty refunds and cost consulting, asset and material
        procurement, equipment maintenance, and official or visitor reception;
  \item Completeness of reimbursement materials (rule 3);
  \item Invoice compliance (rule 4);
  \item Advance-loan eligibility (rule 20); and
  \item Reimbursable-amount calculation (rules 7, 10, and 13).
\end{enumerate}

\paragraph{Step 2: Authoring case templates.}
Using the formal rule base, GPT-5.3 assists in constructing HOL templates for
questions such as whether an expense is reimbursable or how much may be
reimbursed. Assigning concrete personnel and case attributes to template
variables generates individual cases at scale while retaining their formal
input structure.

\paragraph{Step 3: Deriving gold answers.}
We evaluate each instantiated template against the rule base in Isabelle/HOL.
The resulting gold answer is supported by a formal derivation rather than a
human-assigned label, providing a checkable target for automated evaluation.

\paragraph{Step 4: Rendering cases in natural language.}
GPT-5.3 translates the instantiated HOL cases into descriptions for human
readers and evaluated models. The formal representation serves as the reference
for checking whether the descriptions preserve the specified field values
and conditions.

\paragraph{Step 5: Constructing compositional cases.}
We increase reasoning demands along two dimensions.

\textbf{Parallel across scopes.}
A case includes 2--5 distinct personnel categories within one reimbursement
scenario. Under rule 7, for example, personnel identity determines the applicable
travel tier, transportation and accommodation standards, and allowances.
The model must therefore maintain several instances of the chain
\emph{identity $\rightarrow$ scope $\rightarrow$ sub-rule $\rightarrow$ local
conclusion $\rightarrow$ result aggregation}. Increasing the number of categories
increases the number of policy-specific reviews, rather than merely repeating
facts under a single shared standard.

\textbf{Composition across rules.}
A case combines 2--5 distinct reimbursement scenarios. The model must identify
each expense type, select its rules, derive local decisions, and aggregate them
into a comprehensive conclusion. Unlike parallel reviews within one scenario,
these components invoke heterogeneous conditions and sub-rule constraints.
They require concurrent chains of \emph{scenario identification $\rightarrow$
rule matching $\rightarrow$ local conclusion $\rightarrow$ result aggregation},
testing rule switching, parallel processing, and integration of multiple subtasks.

\subsubsection{Prompt Design and Quality Control}

\paragraph{Prompt design.}
Each task-specific prompt specifies the allowed information sources, reasoning
requirements, structured output schema, decision criteria, and a final format
check. The resulting machine-parsable audit report can be compared automatically
with the solver-derived answer. Task-specific criteria disambiguate different
output semantics rather than imposing one decision format on every question.

\paragraph{Quality control.}
Automatic checks compare natural-language descriptions with their formal HOL
field values. Manual inspection of model outputs identifies recurring errors
that motivate prompt refinements; for example, explicit guidance distinguishes
missing required materials from genuinely ambiguous cases. We validate separate
evaluation templates for material completeness, invoice compliance, reimbursement
approval, labor fees, amount calculation, and loan eligibility. Cases whose
wording inadequately conveys the formal conditions are filtered out. These checks address both faithful
case presentation and alignment between each task's decision semantics and
its evaluation format.

\begingroup

\subsection{HOL Reasoning Type Annotation}
\label{app:hol-type-annotation}

\subsubsection{Annotation Scope and Evidence}

\paragraph{Scope and annotation unit.}
We assign the five types in Table~\ref{tab:hol-taxonomy} at the case level. The annotation covers 1{,}079 distinct finance cases and 300 law cases from ten rule bundles. Original and enriched records, archived outputs, and trace copies are not counted as additional cases. For a case $x$, its label set $L(x)\subseteq\{\mathrm{PR},\mathrm{PT},\mathrm{PC},\mathrm{PS},\mathrm{PQ}\}$ may contain multiple types or be empty. The labels describe the policy operations involved in the complete task, rather than only the operations visited along one short-circuit evaluation path.

\paragraph{Evidence and annotation records.}
Finance annotations use the reimbursement-rule definitions and the case-specific formal input conditions. Law annotations use the rule catalog, gold trace, embedded proof, and conflict definitions. Each record stores a label list, five Boolean flags, an annotation version, and evidence linking each assigned type to the relevant source definitions and conditions. Finance evidence additionally identifies the component case; law evidence includes the relevant articles, rule numbers, and source files. 

\subsubsection{Type-Specific Criteria}

\paragraph{Policy Relation (PR).}
PR requires a case-relevant relation between policies. In finance, a contract-specific transport policy must actually be activated to count as overriding the ordinary transport policy. In law, both endpoints of a priority or exception relation must be triggered. A relation present in the rule base but inactive in the case is insufficient.

\paragraph{Policy Transformation (PT).}
PT covers case-triggered modifications of a base policy, including limit adjustments, restrictions of applicability, and exception review. Finance examples include reduced allowances after a duration threshold and exceptions to documentation requirements; reviewing an exception does not imply its eventual approval. In law, an activated exception excludes a base article through the defeat and applicability definitions. Ordinary priority alone does not additionally receive PT. At the semantic level, a fixed exception predicate $E$ restricts a base policy $P$ as $T_E(P)(z)=P(z)\land\neg E(z)$.

\paragraph{Policy Composition (PC).}
PC requires combining separately defined policy predicates into a joint review policy or applying explicitly linked policy constraints. Conjunctions of ordinary facts, intermediate factual concepts, or independently triggered articles do not suffice. Finance includes named subpolicy combinations and activated cross-rule links. Under this criterion, the current law cases receive no PC labels.

\paragraph{Policy Selection (PS).}
PS requires a contextual distinction that selects an applicable policy branch, such as expense type, role, funding source, location, or payment stage. Selecting a fixed role- or region-dependent standard counts as PS rather than PT. Merely belonging to an expense category does not establish a case-specific policy-selection operation. In law, a Boolean applicability test and a final legal consequence are not treated as an independent operation returning a policy; consequently, the current law cases receive no PS labels.

\paragraph{Policy Quantification (PQ).}
PQ requires quantification over policies or a policy domain, including existential quantification and its negation. The current finance cases contain no such operation. Law uses a finite article domain to check for overriding, exempting, or conclusion-producing articles. These quantified final-result definitions are used even when only one article is triggered, so all law cases receive PQ.

\subsubsection{Composite Cases and Validation}
\paragraph{Composite cases and exclusions.}
For finance cases with multiple components, labels are combined by set union. Independent batch review does not by itself create PC or PQ. Nested rule records contribute labels only when their linking conditions are active; unused default records are excluded. Twenty standalone invoice-completeness cases have no matching policy-level operation and retain empty label sets. Thus 1{,}359 of the 1{,}379 cases carry at least one type label. 

\paragraph{Reproducibility and interpretation.}
Domain-specific annotation procedures recompute the labels, supporting evidence, and summary statistics and compare them with the stored annotations. Finance checks original/enriched case identities and rejects unknown condition structures. Law checks triggered relations, defeated and applicable articles against the gold trace, and checks the presence of source relations, quantified definitions, and referenced conclusion definitions. These checks do not rerun the Isabelle prover. The labels are reproducible policy-semantic annotations, not an independently human-adjudicated gold standard. The implementation may encode the operations using record predicates, branches, or a finite article datatype: for example, an article $a$ denotes the policy predicate $P_a(z)=\operatorname{Triggered}(a,z)$. The labels do not assert that every example implements the higher-order function signatures in Table~\ref{tab:hol-taxonomy}.

\endgroup

\subsection{Dataset Illustrations}
\label{app:dataset-examples}
\begin{figure*}[ht]
    \centering
    % \fbox{
    % \begin{minipage}{0.95\textwidth}
        \centering
        % \vspace{3em}
        \includegraphics[width=\linewidth]{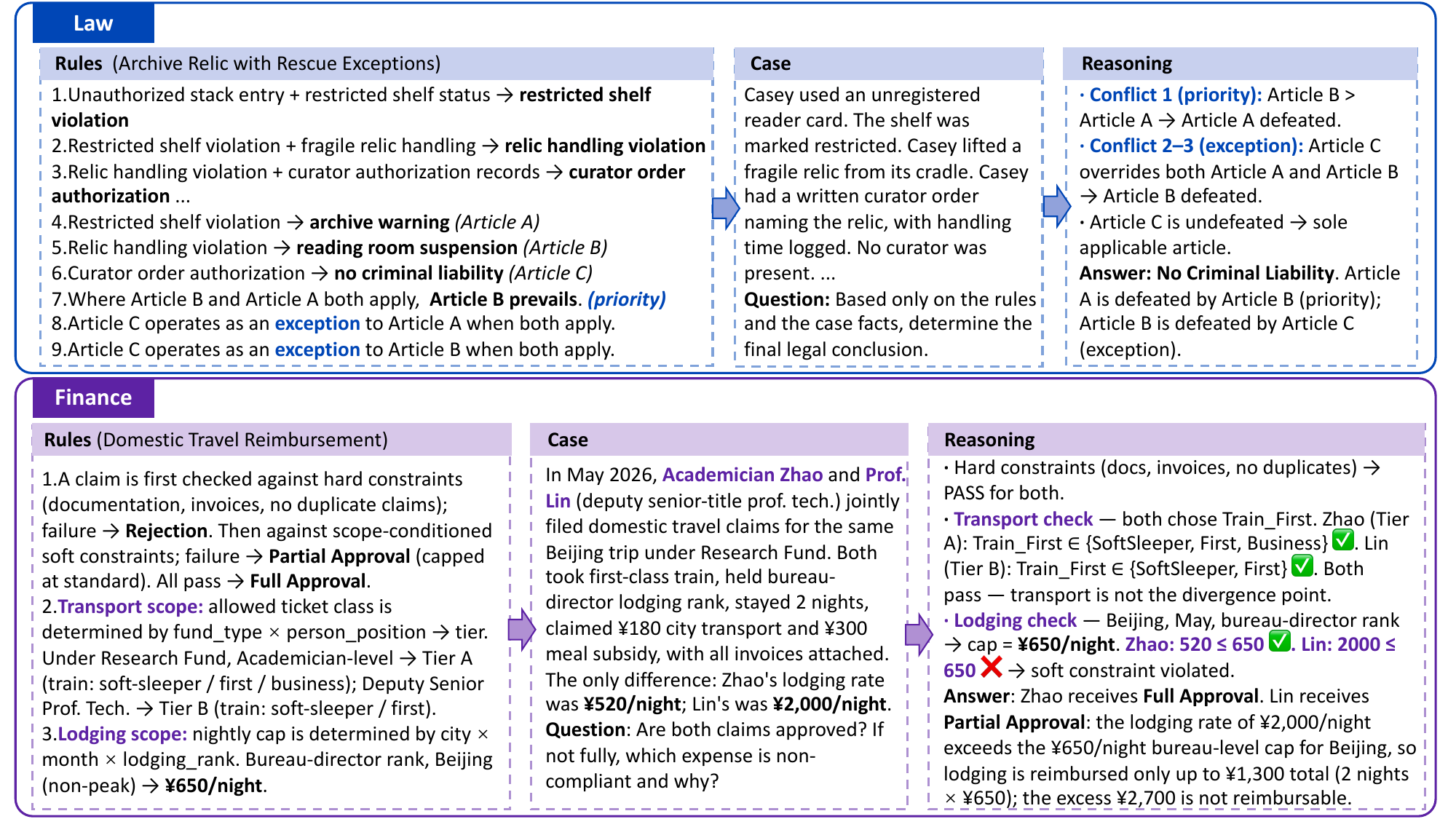}
    \caption{
    Example instances from \textsc{HOLogicBench}, illustrating the higher-order structures in the law and finance portions of the dataset.
    }\label{fig:HOLogicBench_examples_text}
\end{figure*}

Figure~\ref{fig:HOLogicBench_examples_text} presents representative instances from \textsc{HOLogicBench}, showcasing the higher-order logical structures that characterize both the law and finance portions of the dataset.

\section{Reference Derivations and Formal Guarantees}
\label{app:reference-details}

\subsection{Soundness and Completeness of Enumeration}
\label{app:tree-enumeration}

We prove Theorem~\ref{thm:tree-enumeration} using the task notation
$x=(\mathcal{R},\mathcal{F},q)$,
$\hat{x}=(\hat{\mathcal{R}},\hat{\mathcal{F}},\hat{q})$, and $a^\star$.
A derivation tree explains the verified answer to $q$: for a Boolean query,
its root asserts $\hat{q}$ or $\neg\hat{q}$ as appropriate; for a decision or
numeric query, it asserts the verified result value.
{These symbolic trees are distinct from the model-generated sequence
$\tau$ and from a rendered reference trace $\mathcal{G}$.
The reference family $\mathfrak{G}_x=\{\rho_x(t):t\in\mathcal{D}_{\min}(x)\}$ is defined in Appendix~\ref{app:trace-rendering},
and the independently maximized multi-reference metrics are defined in Section~\ref{sec:reasoning-trace-evaluation}.
The theorem concerns the symbolic enumeration stage; it does not establish
that text-alignment scores certify a model-generated proof.}

\subsubsection{Explanation Specification and Tree Identity}
\label{app:tree-specification}

Fix an instance $x$. Its explanation specification consists of a finite node
set $V_x$, root $v_x^\star$, labels $\lambda_x(v)$, and finite sets of
alternative premise lists $\mathcal{J}_x(v)$.
A label records the node kind, expression, and value, including the named
business rule when applicable. An alternative
$\alpha=(v_1,\ldots,v_k)\in\mathcal{J}_x(v)$ requires all listed premises;
different alternatives are different sufficient justifications.
The empty list is an admissible leaf, whereas an empty set of alternatives
admits no derivation. A height function $h_x:V_x\to\mathbb{N}$ satisfies
$h_x(v_i)<h_x(v)$ for every premise of every alternative of $v$.
This is the finite acyclicity assumption, not a restriction on all HOL theories.

\paragraph{Local reasoning semantics.}
A true conjunction requires both true operands; a false conjunction admits
each false operand independently. A true disjunction admits each true operand
independently; a false disjunction requires both false operands.
A true implication admits a false antecedent or a true consequent whenever
that reason holds; a false implication requires a true antecedent and a false
consequent. An ordered conditional requires its guard proof and only the
selected branch. Thus later branches retain the evidence excluding earlier
branches. Rule unfolding, case analysis, and deterministic computations use
their declared local justifications. Numeric lookups may be atomic computations
at the chosen explanation granularity; internal table scans are not additional
business derivations.

\paragraph{Canonical trees.}
Let $C_x(v;t_1,\ldots,t_k)$ construct a node labeled $\lambda_x(v)$ with
recursively canonicalized children, removing identical children and sorting
the remainder in a fixed total order. Trees differing only in premise order
or duplicate identical premises are identified. Distinct labels and distinct
remaining subtree structures are preserved. This normalization concerns proof
premises, not operand order inside expressions: it does not make implication,
subtraction, or conditional branches commutative. Semantically relevant roles
must remain identifiable; an identical proof may discharge repeated obligations.

Define $\mathcal{D}_x(v)$ as the least family of canonical trees closed under
the following rule, including $k=0$:
\begin{equation}
\frac{\begin{gathered}
 (v_1,\ldots,v_k)\in\mathcal{J}_x(v)\\
 t_i\in\mathcal{D}_x(v_i)\quad(1\leq i\leq k)
\end{gathered}}
{C_x(v;t_1,\ldots,t_k)\in\mathcal{D}_x(v)}.
\label{eq:tree-admissibility}
\end{equation}

\paragraph{Faithfulness assumption.}
At the declared explanation granularity, $\mathcal{J}_x(v)$ contains exactly
the permitted local business justifications. Leaves are valid under
$\hat{\mathcal{F}}$, and each local justification preserves validity under
$\hat{\mathcal{R}}$ when its premises are valid. Source extraction and
canonicalization preserve these obligations, priorities, and computation
boundaries. This assumption explicitly separates faithful source translation
from the enumeration theorem; evaluating a root answer alone does not establish it.

\subsubsection{Exhaustive Enumeration}

For each alternative $\alpha=(v_1,\ldots,v_k)$, define
\begin{equation}
\begin{split}
\mathcal{U}_x(v,\alpha)=\{\,&C_x(v;t_1,\ldots,t_k):\\
 &(t_1,\ldots,t_k)\in\textstyle\prod_{i=1}^k
                  \operatorname{Enum}_x(v_i)\,\},
\end{split}
\end{equation}
and recursively enumerate
\begin{equation}
\operatorname{Enum}_x(v)=
\bigcup_{\alpha\in\mathcal{J}_x(v)}\mathcal{U}_x(v,\alpha).
\label{eq:tree-enumeration}
\end{equation}
The empty Cartesian product contains the empty tuple. Intermediate results
are memoized and structurally deduplicated, but are not pruned by evidence or size.

\begin{lemma}[Exact enumeration]
\label{lem:tree-exact}
For every $v\in V_x$, enumeration terminates and
$\operatorname{Enum}_x(v)=\mathcal{D}_x(v)$.
\end{lemma}
\begin{proof}
Induct on $h_x(v)$. Recursive calls have strictly smaller height. Since each
alternative set and premise list is finite, induction gives finite premise
sets and finite Cartesian products. Thus enumeration terminates, including
the leaf case represented by the empty product.

For soundness, each generated tree combines premise trees from an admissible
alternative. By induction those premise trees belong to
$\mathcal{D}_x(v_i)$; Eq.~\ref{eq:tree-admissibility} admits their canonical parent.

For completeness, take $t\in\mathcal{D}_x(v)$. Its last inductive step supplies
an alternative $(v_1,\ldots,v_k)$ and trees $t_i\in\mathcal{D}_x(v_i)$ such
that $t=C_x(v;t_1,\ldots,t_k)$. By induction each $t_i$ is enumerated.
Their tuple is therefore included in the Cartesian product, and $t$ is
generated. Canonical deduplication removes only repeated representations of
the same tree, never a distinct canonical tree.
\end{proof}

\subsubsection{Global Minimality and Proof of the Main Theorem}

For a canonical tree $t$, write $\operatorname{Ev}_x(t)$ for its set of input
evidence pairs (expression, value), $\operatorname{Ru}_x(t)$ for its set of
named business rules, and $\lVert t\rVert$ for its number of nodes when
recursively unfolded. Shared storage does not reduce this tree size.
Define strict dominance $\triangleleft_x$, distinct from the legal-rule
priority relation $\prec$, by
\begin{equation}
\begin{split}
u\triangleleft_x t\ \Longleftrightarrow\;&
\operatorname{Ru}_x(u)=\operatorname{Ru}_x(t)\\
&\land\bigl[\operatorname{Ev}_x(u)\subsetneq\operatorname{Ev}_x(t)\\
&\quad\lor\bigl(\operatorname{Ev}_x(u)=\operatorname{Ev}_x(t)\\
&\qquad\land\lVert u\rVert<\lVert t\rVert\bigr)\bigr].
\end{split}
\label{eq:tree-dominance}
\end{equation}
For any finite tree set $S$, let
\begin{equation}
\operatorname{Min}_x(S)=
\{t\in S:\nexists u\in S,\ u\triangleleft_x t\}.
\end{equation}
In particular,
\begin{align}
\mathcal{D}_{\min}(x)&=\operatorname{Min}_x(\mathcal{D}_x(v_x^\star)),\\
\operatorname{EnumMin}(\hat{x})&=
\operatorname{Min}_x(\operatorname{Enum}_x(v_x^\star)).
\end{align}
Different named-rule sets are deliberately incomparable. Structurally distinct
trees with equal evidence, rule sets, and size are all retained. This is the
specified business minimality criterion, not unrestricted logical minimization.

\begin{proof}[Proof of Theorem~\ref{thm:tree-enumeration}]
Lemma~\ref{lem:tree-exact} establishes termination and equality of the raw
enumerated and admissible tree sets. Applying $\operatorname{Min}_x$ to both
sides yields
$\operatorname{EnumMin}(\hat{x})=\mathcal{D}_{\min}(x)$.
In particular, an admissible non-dominated tree is enumerated and has no
enumerated dominator, so it cannot be removed. Conversely, every retained tree
is admissible and has no dominator in the complete candidate set.
Canonical set identity ensures that each tree occurs exactly once.

For semantic correctness, induct over each admissible derivation: its leaves
are valid for the case, and each local rule preserves validity by the
faithfulness assumption. Hence its root establishes $a^\star$ for $\hat{q}$
under $\hat{\mathcal{R}}$ and $\hat{\mathcal{F}}$. Ordered branch rules retain
the guard evidence and exclude inactive branches, preserving branch priority.
\end{proof}

\paragraph{Root-only streaming filter.}
The implementation need not store every root candidate simultaneously.
Maintain a set $K$ of non-dominated complete root trees. Reject a new tree if
an element of $K$ dominates it; otherwise remove the elements it dominates
and insert it, suppressing structural duplicates.

\begin{lemma}[Streaming-filter equivalence]
After processing any finite sequence of root candidates, $K$ equals
$\operatorname{Min}_x(S)$ for the set $S$ of candidates processed so far.
\end{lemma}
\begin{proof}
The relation $\triangleleft_x$ is irreflexive and transitive. Rule-set equality
is preserved; proper evidence inclusion composes, and when evidence remains
equal, strict size inequalities compose. In a finite set, every nonminimal
tree is therefore dominated by a minimal tree, by repeated descent and
transitivity. Induct on the candidate sequence. Comparing a new tree against
$K$ detects exactly whether an old candidate dominates it. A dominated new
tree cannot dominate an old minimal tree, by transitivity. Otherwise inserting
it and deleting exactly the old elements it dominates gives the new minimal
set. Duplicate suppression does not affect this invariant.
\end{proof}

\paragraph{Why intermediate pruning is excluded.}
Let $A=B=\mathrm{True}$ and set
$Z=((A\land\mathrm{True})\land\mathrm{True})\lor(A\land B)$, with target
$Z\land B$. At $Z$, the left justification uses $\{A\}$ and the right uses
$\{A,B\}$; local evidence pruning would discard the right one.
At the root, both use $\{A,B\}$ because the sibling already requires $B$.
Counting literal and connective nodes, the complete trees have sizes eight
and six, respectively. The discarded justification is needed for the global
minimum. Dominance is thus not preserved by arbitrary tree contexts, which is
why minimization is applied only after forming complete root trees.

\paragraph{Scope of the guarantee.}
The result assumes faithful local alternatives, the stated tree identity,
finite acyclicity, and an untruncated run. It is not an enumeration of all
Isabelle/HOL proofs, nor an independent proof of source extraction or
serialization correctness. Unsupported constructs and resource limits must
be reported as incomplete. The theorem also does not imply that omission from
a model's natural-language trace establishes omission from its internal reasoning.

\begingroup

\subsection{Rendering Reference Traces}
\label{app:trace-rendering}

\paragraph{Finance: dependency-ordered step templates.}
The renderer reads the exported explanation graph, including node kinds, expressions, evaluated values, premise links, and complete-path roots. For each root, a depth-first postorder traversal visits premises before their conclusions and emits each reachable node once, assigning consecutive step numbers within that path. Shared premises are referenced by their step numbers. Node-specific templates distinguish evidence, Boolean inference, negation, named-rule evaluation, conditional selection, and other computations. For example, a false input fact is rendered as ``Evidence: has signed contract (input 0) = False.'' An inference states its operation and result and cites its supporting steps. Conditional templates state the guard, selected branch, and result. Isabelle symbols are mapped to readable operators such as $\land$, $\lor$, and $\rightarrow$, and underscores in identifiers are replaced with spaces. Within inference expressions, already evaluated subexpressions are replaced with step references to avoid repeated expansion; the exported catalog retains the original expressions, node identifiers, premise links, and input bindings.

\paragraph{Law: signed facts and rule-status statements.}
Each reference begins with the selected factual support, represented as predicate--truth-value pairs. Facts are verbalized using positive or negative sentences from the rule-bundle metadata. If a negative template is missing or duplicates the positive template, the renderer explicitly negates the positive sentence; a readable predicate name with ``holds'' or ``does not hold'' is the fallback. The renderer then emits the named claims included in that explanation in dependency order. Templates state whether an intermediate condition is established and whether an article is triggered, defeated, or applicable, using article titles from the metadata when available. Intermediate conditions and article-triggering statements carry source rule numbers where mapped. The trace ends with the case decision. This is a rendering of named business-level claims, rather than a verbalization of every low-level Boolean connective.

\paragraph{Alternative paths and composite cases.}
Finance retains a separate reference record for every exported path; identical step strings share storage without merging paths. For a composite case, one complete path is chosen for each component, and the rendered blocks are concatenated under scenario headings. The reference family contains the Cartesian product of these component choices. Law removes only identical rendered texts, retaining a reference identifier, signed support, and validation-lemma identifier for each retained record. Both pipelines use fixed templates at this stage rather than an LLM paraphrasing pass. Thus the evaluation compares generated reasoning with reproducible textual renderings of symbolic explanations; rendering itself is not covered by the enumeration theorem.

\endgroup

\begingroup

\section{Main Evaluation Implementation}
\label{app:reasoning-implementation}

The following details describe the implementation of the metrics in
Section~\ref{sec:reasoning-trace-evaluation} for Table~\ref{tab:main_results}.

\subsection{Model Configurations}
\label{app:evaluation-data}

\paragraph{Model variants and reasoning modes.}
Table~\ref{tab:main_results} lists the evaluated models and variants.
Thinking comparisons use Qwen3-30B Instruct and Thinking variants,
MiniMax-M3 with adaptive or disabled thinking. Each configuration in these
paired comparisons covers all 1,379 benchmark cases. The model reported
as Gemini-3.1-Flash uses Gemini-3.1-Flash-Lite.

\paragraph{Output limits and evaluation inputs.}
DeepSeek-R1 requests use completion limits of 16,000 or 32,000 tokens.
Evaluation uses the final answer and the accompanying reasoning trace;
empty responses and missing traces are handled by the aggregation protocol
below. The counterfactual experiment uses a separate generation setup,
detailed in Appendix~\ref{app:counterfactual-implementation}.

\subsection{Reference Handling and Score Aggregation}
\label{app:evaluation-aggregation}

We apply the independent per-metric maximization and case averaging in
Eq.~\ref{eq:multireference-score}. References retain distinct factual evidence
and Boolean branches. Overlapping HOL categories do not duplicate cases in
the Overall score. Empty responses and missing reasoning traces receive zero
reasoning scores and remain in the averaging denominator.

\subsection{Similarity Metrics}
\label{app:similarity-metrics}

\paragraph{BERTScore implementation.}
We use \texttt{ModernBERT-base} representations from layer 22.
Long texts are encoded in 8,192-token windows with 1,024-token overlap
(up to 8,190 content tokens plus special tokens per window).
Overlap tokens count once: the overlap is divided at its midpoint to assign
each original content token to a single window representation.
Precision and recall use global token matching across the full candidate and
reference texts, followed by their harmonic mean; we do not average
window-level F1 scores or restrict matching to corresponding windows.

\paragraph{ROSCOE-SA-P implementation.}
We use normalized \texttt{all-mpnet-base-v2} sentence embeddings with a
512-token limit per step.

% \paragraph{Rule-ID matching.}
% Finance matches main section IDs, including structured selections; subclauses
% such as 7.2 and 7.3 both map to section 7 before deduplication.
% Law uses source rule IDs, including evaluated priority rules.
% Reference sets include positive and negative rule checks.
% Repeated IDs count once; paraphrases are not matched.

\endgroup

\begingroup

\section{Implementation Details of Analysis Experiments}
\label{app:analysis-implementation}

\subsection{Shared Data and Reporting Protocol}
\label{app:analysis-protocol}

The analyses use the case identities, formal references, and saved model outputs
underlying the main evaluation. We distinguish a source case, a model--case
output, and a pair of outputs for the same case under two model configurations.
Full-benchmark depth and thinking analyses use all 1,379 cases. HOL types overlap:
a case contributes to each of its assigned types but only once to Overall.
The 20 cases with no HOL type contribute only to Overall. In contrast, the
trace, intervention, and error-analysis samples use a designated sampling type
for each selected unit, while retaining its original overlapping labels.
Their balanced or quota-based distributions therefore differ from that of the
full benchmark. Sampling manifests preserve case identities and seeds; saved
predictions, reference identifiers, and validation records determine the reported
results. No new benchmark answers are generated for the depth, thinking, or
error analyses. The intervention experiment regenerates all three versions of
each selected case under its own shared prompt.

\subsection{Derivation-Level Trace Evaluation}
\label{app:derivation-analysis}

\paragraph{Sample and candidate references.}
We select 500 distinct cases with seed 20261010, assigning 100 to each HOL
type. Types are processed from the smallest eligible population to the largest
(PT, PR, PQ, PS, PC); each draw is uniform without replacement among the
remaining eligible cases. The same cases are used for all five models in
Table~\ref{tab:reasoning}, giving 2,500 model--case outputs. For each output,
the retrieval stage ranks the case's exported reference paths by semantic
step alignment and retains up to three paths. Retrieval uses nonblank original
answer units except standalone bracketed headings; factual units are retained
at this stage. Reference units are canonical derivation texts, including the
full expression and evaluated value for Finance and named derivations for Law.
We use normalized \texttt{all-mpnet-base-v2} embeddings, with long units
split into token chunks rather than truncated. A step pair takes the maximum
chunk-to-chunk cosine similarity; bidirectional mean-max aggregation then
yields retrieval precision and coverage, combined into a retrieval F1.
Steps receive equal weight regardless of their number of chunks, and retrieval
ties are resolved by reference identifier. Composite Finance paths are ranked
over the complete Cartesian family using component-level maximum/sum factoring.
The frozen selection yields 5,120 candidate-path judgments for the 2,500
outputs. The retrieved subset bounds the alternatives considered by the judge;
it is not an exhaustive search for the best judge-scored reference.

\paragraph{Fixed units and critical steps.}
Generated-step identifiers and their original text are fixed before alignment
and reused across the candidate references. Reference derivation identities
preserve operation kind, expression or named definition, evaluated value, and
component/case identity. Facts, literals, and constructors provide context
rather than reference derivations; negative checks and the final derivation
remain in the reference denominator. Critical steps are defined before
retrieval as the intersection of canonical derivation identities across
\emph{all} exported references for that case. They therefore include common
logical and computational operations as well as policy-level operations.
They are not restricted to the five HOL type labels or selected separately
for each model. This operational definition concerns the finite reference
family, not necessity in every possible HOL proof.

\paragraph{Trace-only judging.}
The final trace-only protocol gives GPT-6-Luna the generated trace and one
trusted reference path, including its selected premise and dependency
information; it does not provide the original case narrative or complete
policy document. Compact local identifiers and a shared string dictionary
reduce repeated text. The judge must use only the supplied traces. Each
generated reasoning unit receives one status: matched, valid but
non-reference, incorrect, or unsupported. Each reference derivation receives
correct, incorrect, unsupported, or not covered, together with the relevant
generated identifiers. A matching conclusion or rule number alone does not
establish a match. Compressed steps may cover multiple reference operations
only when they justify each operation; multiple nonredundant generated steps
may jointly establish one reference operation. A mixed unit with an incorrect
or unsupported necessary part receives no partial alignment credit.
Correct extra reasoning outside the selected path is distinguished from an
error, but is credited as valid only when justified by the supplied traces.
The final compact response contains statuses and identifier links rather than
free-form rationales, quoted evidence, or numeric scores.

\paragraph{Scoring and reference selection.}
For one candidate reference, let $G$ be the number of generated reasoning
units, $M$ the number marked matched, $R$ the number of reference derivations,
and $C$ the number correctly covered. Then
\begin{align*}
P &= M/G,\qquad \mathrm{Coverage} = C/R,\\
\mathrm{Deriv\mbox{-}F1} &= \frac{2P\,\mathrm{Coverage}}
                                  {P+\mathrm{Coverage}}.
\end{align*}
Precision is zero when there are no generated reasoning units, and F1 is
zero when precision and coverage are both zero.
Critical-step Correctness is the fraction of the precomputed critical
derivations marked correct, under the same alignment standard.
The scorer computes these quantities locally. All four metrics for a
model--case output come from the reference with the highest derivation F1;
there is no independent maximization of precision, coverage, or critical
correctness. F1 ties are resolved by retrieval rank and then reference
identifier. A case with no critical derivations would have an undefined
critical score and be excluded from that metric's mean; the exported
benchmark has no such cases. Results are then macro-averaged over the 500 cases and multiplied
by 100, rather than pooling steps across cases.

\paragraph{Validation and retry policy.}
The initial run uses reasoning effort \texttt{high} and a 32,768-token output
ceiling. Subsequent rounds use \texttt{max} for failed candidate judgments,
retaining already valid judgments. Validation checks structured-output
completeness, identifier coverage, allowed statuses, and bidirectional
alignment: a generated match and a correctly covered reference must link to
each other consistently. Invalid or incomplete judgments are retried rather
than repaired by inserting inferred alignments. A winning reference is
selected only after every retrieved candidate for that output has a valid
judgment. The final collection contains all 2,500 scored outputs and no
unresolved candidate requests. These checks establish structural consistency,
not independent agreement between two judges; the reported evaluation uses
one judge model with selective retries. Its trace-only access also limits
assessment of alternative reasoning not justified by either supplied trace.

\subsection{Counterfactual Intervention Protocol}
\label{app:counterfactual-implementation}

\paragraph{Eligibility and sampling.}
The intervention changes exactly one factual input field while keeping the
rule base fixed. A case is eligible only if it admits both a decisive edit
that changes its gold decision and a control edit that preserves it.
The decisive field must occur in the intersection of the factual support sets
of all exported references for the original case. Using seed 20261009, we
select 250 distinct originals and remove duplicate formal-input signatures.
Each case is assigned to one sampling stratum: PR and PQ each contain 50 Law
cases, PC and PS each contain 50 Finance cases, and PT contains 38 Law and
12 Finance cases. Selection cycles through buckets defined by source,
dependency depth, and original gold answer. This gives 138 Law and 112 Finance
cases. A decisive edit is sampled from the eligible candidates; the control
preferentially uses the same value type as the decisive edit.

\paragraph{Construction and formal checks.}
For Law, we flip one Boolean fact, evaluate the unchanged rule definitions,
and retain only variants with a unique final consequence. For Finance, we
match existing formal input records with the same evaluator that differ in
exactly one field; all other record bindings must agree. In a composite case,
only one component is changed, and the gold answer remains an ordered vector
of component decisions. The saved verification record reports 414 Law variant
theorems and 173 Finance component equalities checked in Isabelle, together
with checks that all 250 decisive edits change the answer and all 250 controls
preserve it. The prompt audit also checks one visible factual edit per variant.

\paragraph{Prompts and inference.}
All three versions are rendered from their formal facts using a common
template and submitted independently. Prompts contain the unchanged rules,
the variant's facts, allowed decision codes, and an output schema with
a reasoning trace and final decisions; they reveal neither
the intervention type nor the gold answer. Thus original-case accuracy in
Table~\ref{tab:counterfactual} is measured on these regenerated originals,
rather than copied from the main table. MiniMax-M3 uses its official endpoint
with \texttt{adaptive} or \texttt{disabled} thinking, temperature 1.0,
a nucleus-sampling probability of 0.95, and a 32,000-token completion limit. M3-T denotes the
adaptive setting, which does not force thinking on every request.
GLM-5.3-Flash uses temperature 0, seed 20261009, and a 32,000-token limit;
its saved responses come from multiple OpenRouter providers. Successful
responses are retained while unresolved requests are retried. Of the 2,250
model--variant requests, 2,248 have valid final responses; two GLM responses
remain incomplete at the length limit.

\paragraph{Metrics and denominators.}
Let $o_i,d_i,c_i\in\{0,1\}$ indicate correct original, decisive, and control
answers, respectively. Correctness requires exact equality of the complete,
ordered decision vector. For $N=250$, the reported proportions are
\begin{align*}
\mathrm{O\mbox{-}Acc} &= N^{-1}\sum_i o_i,\\
\mathrm{CF\mbox{-}Acc} &= N^{-1}\sum_i d_i,\\
\mathrm{CRS} &= \frac{\sum_i o_i d_i}{\sum_i o_i},\\
\mathrm{All\mbox{-}3} &= N^{-1}\sum_i o_i d_i c_i.
\end{align*}
Ctrl-Fail is the fraction of cases for which the control prediction differs
from the original prediction, or either prediction is unavailable. It does
not require the original prediction to be correct. Missing or invalid answers
count as incorrect for accuracy and All-3 and remain in the unconditional
denominator. CRS conditions on each model's own correctly answered originals,
so its denominator differs across models. Type-specific results use the
50 cases assigned to that stratum. The scoring script also records Wilson
95\% intervals and the exact numerator and denominator of each rate. These
interventions measure behavioral sensitivity to facts under fixed rules;
they do not directly identify a model's internal reasoning process.

\subsection{Reasoning Depth Analysis}
\label{app:depth-analysis}

\paragraph{Depth extraction.}
Depth is the maximum number of named derived-rule applications along a
fact-to-target dependency chain in the exported explanation system.
Input facts have depth zero. Boolean connectives, arithmetic operations,
and administrative wrappers do not themselves add a policy-rule layer.
For Finance, we count named definitions in the reimbursement rule base, including
named helper, selection, and aggregation definitions, and exclude case-harness
wrappers. We take the maximum over reference roots and then over the components
of a composite case. Parallel premises likewise combine by maximum, not sum.
For Law, we use the signed facts and Boolean definitions employed in reference
generation, count named intermediate concepts, triggers, defeat and
applicability checks, and follow branches supporting the evaluated outcome.
The no-liability definition supplies the root for no-liability cases; other
outcomes additionally count the final legal-result definition.
This quantity depends on the granularity of the formalization and is neither
the shortest sufficient proof depth nor the longest unrestricted HOL proof.

\paragraph{Aggregation and outcome stratification.}
We first average binary answer correctness over the 16 model configurations
for each case and then average cases within each exact depth and HOL type.
Each depth group retains its case count. The supporting analysis uses 2,000
case-bootstrap replicates with seed 20261009 for 95\% intervals and also
checks sensitivity to removing depth groups containing fewer than 10 cases.
Figure~\ref{fig:depth_accuracy} displays the group means. Within PQ, the
two subgroups are defined by the gold outcome: no criminal liability
versus every other outcome. This is an
outcome-based split, not a direct indicator of whether an exemption rule is
applicable. The curves describe associations within these groups; they do
not isolate a causal effect of depth or exemptions.

\subsection{Paired Thinking Analysis}
\label{app:thinking-implementation}

\paragraph{Full-benchmark pairing.}
We join the thinking and non-thinking or instruct outputs by case identity
for DeepSeek-V4-Flash, Qwen3-30B, and MiniMax-M3. Each configuration contains
the same 1,379 cases. For a group of $n$ cases, let $n_{01}$ count
non-thinking-incorrect/thinking-correct pairs and $n_{10}$ the reverse.
The reported accuracy difference in percentage points is
$100(n_{01}-n_{10})/n$. We also retain counts for both-correct and both-wrong
pairs. These comparisons have different operational meanings: Qwen compares
separately named Instruct and Thinking variants; MiniMax compares adaptive
and disabled thinking; V4-Flash compares enabled and disabled thinking.

\paragraph{Selection of paired examples.}
The qualitative analysis contains 52 pairs: 12 previously selected Qwen
pairs, 20 MiniMax pairs, and 20 V4-Flash pairs, comprising 104 outputs from
44 distinct source questions. The latter two samples each select two gains
and two losses per designated HOL type from the 254 source questions in
the error-analysis dossiers. Candidates are ordered deterministically by
a SHA-256 key incorporating seed 20261010, model family, type, and case
identity. Types are processed in the order PC, PS, PR, PT, PQ, and questions
are not reused within a model family. This is a targeted sample of changes
in correctness, not a random sample of all paired outputs.

\paragraph{Mechanism assessment.}
GPT-6 Astra compares the two outputs with the case and formal references,
assigning one of four verdicts: verified improvement, verified regression,
evaluation or specification issue, or mixed/uncertain. Mechanism labels
cover constraint application, branch selection, object binding, numerical
computation, exception/priority handling, unsupported assumptions, and
reasoning--output consistency, and may overlap. The verification script
checks that each quoted passage occurs verbatim in the corresponding
output and that the stated correctness transition matches the saved scores.
The saved analysis contains 17 verified improvements, 5 verified regressions,
14 evaluation/specification issues, and 16 mixed/uncertain pairs. These are
model-assisted assessments of selected examples, rather than a population
estimate of how often thinking helps or harms reasoning.

\subsection{Error Analysis Protocol}
\label{app:error-analysis}

\paragraph{Sampling unit and quotas.}
We sample outputs from Qwen3-30B-Instruct, Qwen3-30B-Thinking, GPT-5.4-Mini,
and GLM-5.3-Flash. Each model/type cell contains 12 originally incorrect
and 4 originally correct predictions, giving 320 distinct model--case pairs
from 254 source questions. We retain the initial 80 sampled outputs and add
9 incorrect and 3 correct outputs per cell using the deterministic seed
key combining seed 20261009, an extension-stage marker, type, model, and correctness.
Previously selected model--case pairs are excluded across overlapping types.
Every output has one designated reporting type. The final sample is therefore
an extension of the initial sample under fixed quotas, rather than an
independent simple random sample of the full prediction population.

\paragraph{Evidence and annotation.}
Each dossier contains the source case, model output, original answer
assessment, and formal reference evidence. Finance dossiers preserve
component input bindings and explanation nodes and paths; Law dossiers
include the valid references and answer-audit record. GPT-6 Astra assigns
an overall audit status and records confirmed error mechanisms. The status
distinguishes a confirmed model error, an ambiguous specification, an
evaluation issue, and no observed substantive error. Confirmed secondary
errors can coexist with specification or evaluation issues, so the absence
of a primary model-error status does not automatically imply sound reasoning.
Checks verify record identities, original correctness labels, and consistency
between confirmed mechanisms and the presence of a substantive error.

\paragraph{Error taxonomy and counting.}
Fine-grained labels are mapped to eight reporting families: fact reading and
unsupported premises; applicability and policy branches; exceptions and
priorities; object and document binding; quota or condition transformations;
numerical computation and aggregation; reasoning--output inconsistency; and
whole-case versus local-decision aggregation. Families are non-exclusive,
but an output counts at most once within a family. The mechanism percentages
use outputs with at least one confirmed error as their denominator: 19 for
PR, 26 for PT, 36 for PC, 44 for PS, and 15 for PQ, out of 64 sampled outputs
per type. Figure~\ref{fig:hol_errors} displays up to three most frequent
families per type; PQ has only two observed families. For PT, reasoning--output
inconsistency and exception/priority handling tie for third, and the former
is displayed. Stacked counts may exceed the number of distinct erroneous
outputs because labels overlap. The 12/80 result uses a separate denominator:
all sampled outputs originally scored as correct.

\paragraph{Review subset and procedure.}
We select 64 outputs (20\% of the annotation sample) independently of
Astra's labels and similarity scores using a deterministic key combining
seed 20261009 with a human-review sampling marker. The subset contains 16 outputs per model,
48 originally incorrect and 16 originally correct predictions, and
13 outputs each for PR/PT/PC/PS and 12 for PQ. A human reviewer
reviews all 64 annotations against the original facts, natural-language
rules, complete model outputs, and supporting formal evidence. Formal
reference graphs are abridged to input bindings, deduplicated evaluated
rule/fact nodes, and final results. The reviewer sees the original annotations
but not the similarity scores, and returns an agreement judgment,
evidence-based justification, and suggested corrections. This is a
non-blind human review of the initial model-assisted annotations.

\paragraph{Human-review instructions.}
The following instructions summarize the review criteria:
\begin{enumerate}[label=(\roman*)]
  \item Read the supplied facts, natural-language rules, complete model
  response, formal evidence, and Astra annotation. Judge the response under
  the supplied rules; do not substitute external legal or reimbursement
  knowledge or treat the existing annotation as authoritative.
  \item Inspect both the reasoning and the final answer, including amounts
  and component-level decisions. In reimbursement cases, treat unlisted
  required documents as missing and evaluate independent expenses or review
  objects separately. A correct final answer does not establish correct reasoning.
  \item Distinguish a definite model error from an ambiguous task
  specification, a reference-answer or scoring issue, and the absence of an
  observed substantive error. If the natural-language task and formal
  encoding differ, record the discrepancy rather than automatically
  attributing it to the model. Mark insufficient evidence as uncertain.
  \item Check the primary classification and each claimed error mechanism
  against specific rules, facts, and response passages. Record independently
  supported secondary errors even when the primary issue concerns the
  specification or scoring.
  \item Select full agreement, partial agreement, disagreement, or uncertainty
  with the annotation. Confirm or correct the primary status and separately
  indicate whether any definite model error is present. Agreement concerns
  the annotation, not whether the underlying answer is correct.
  \item Record confidence and the confirmed or revised mechanisms. For
  partial agreement or disagreement, cite the relevant output and rule or
  fact, explain the discrepancy, and provide a suggested correction before
  marking the review complete.
\end{enumerate}

\paragraph{Review agreement.}
The reviewer fully agrees with 46/64 annotations (71.9\%) and partially
agrees with 18/64 (28.1\%); none receives an overall judgment of disagreement
or uncertainty. Partial agreement can retain the main finding while
suggesting an additional error, a different primary status, or a revised
mechanism. We therefore report exact field agreement separately:
the four-way audit status agrees in 54/64 cases (84.4\%), and the binary
judgment of whether any definite model error is present agrees in
57/64 (89.1\%). Table~\ref{tab:error-review-agreement} breaks these results
down by reporting type. Full agreement is an endorsement of the annotation,
not a claim that the underlying model output is correct; partial agreement
is not counted as a pass.

\begin{table}[t]
\centering
\small
\setlength{\tabcolsep}{4pt}
\begin{tabular}{@{}lrrrr@{}}
\toprule
\textbf{Type} & $n$ & \textbf{Full} & \textbf{Status} & \textbf{Any error} \\
\midrule
PR & 13 & 9 & 10 & 10 \\
PT & 13 & 8 & 10 & 13 \\
PC & 13 & 9 & 12 & 12 \\
PS & 13 & 8 & 10 & 10 \\
PQ & 12 & 12 & 12 & 12 \\
\midrule
All & 64 & 46 & 54 & 57 \\
Rate (\%) & -- & 71.9 & 84.4 & 89.1 \\
\bottomrule
\end{tabular}
\caption{Human review agreement on the 64-output subset.
Full: overall full agreement; Status: exact agreement on the four-way
audit status; Any error: agreement on the presence of at least one
definite model error. Counts use the row-specific $n$ as denominator.}
\label{tab:error-review-agreement}
\end{table}

\paragraph{Interpretation.}
These results measure agreement between a human reviewer and the initial
model-assisted annotations, rather than annotation accuracy or agreement
between independent human annotators. The original annotations are visible
during review. We retain the saved Astra annotations for all reported
mechanism counts rather than automatically incorporating the reviewer's
proposed changes. The deliberately error-enriched sample does not estimate
population error prevalence. Human review covers the 64-output subset and
does not establish a fully human-adjudicated gold standard for all annotations.

\endgroup

\section{Ethics and Use of AI}
\label{app:ethics-ai}

\subsection{Data Sources and Privacy}

\paragraph{Source materials.}
The Law dataset contains five rule bundles inspired by the Chinese Criminal
Law text and five fictional regulatory systems. The former rewrite and
simplify selected provisions into self-contained reasoning tasks; they are
not authoritative representations of current law. The Finance dataset adapts
a publicly accessible reimbursement policy from Fudan University into formal
rules and synthetic cases. Source links are provided in the dataset
construction description. We do not claim ownership of the underlying legal
or institutional texts.

\paragraph{Synthetic cases and sensitive content.}
The benchmark uses constructed case facts rather than real legal case files,
personal financial records, or reimbursement claims. Names and case entities
are synthetic. Legal-style scenarios may concern criminal liability;
descriptions focus on facts needed for rule application and avoid unnecessary
graphic or identity-targeted detail. Author inspection and the formal--text
consistency checks described in Appendix~\ref{app:dataset-details} support
data quality and alignment with the encoded rules and facts.

\subsection{Intended Use and Broader Impact}

\paragraph{Intended use.}
\textsc{HOLogicBench} supports research on higher-order reasoning under supplied
rules and facts. Its simplified legal and reimbursement settings are not
intended to provide legal, financial, or institutional policy advice.
Benchmark performance alone does not establish fitness for real-world
eligibility, liability, or payment decisions.

\paragraph{Benefits and limitations.}
Formal answers and multiple reference derivations enable analysis beyond
final-answer accuracy, helping identify incomplete justifications and failures
in rule application. These benefits are bounded by the encoded rules and the
coverage of the synthetic cases. Formal verification establishes consequences
of an encoding; it does not certify that the encoding captures every real-world
requirement or that a model-generated explanation is valid. Model-assisted
trace judgments and targeted qualitative samples introduce additional
uncertainty and should not be treated as independently verified measures of
real-world reliability.

\subsection{Human Oversight and AI Assistance}

\paragraph{Dataset construction.}
The authors conduct manual inspection and correction during construction and
quality control; no crowdsourced annotation is used. GPT-5.3 assists with
initial HOL formalization and, for Finance, case-template construction and
natural-language rendering. Benchmark answer labels are obtained through
Isabelle/HOL verification of the encoded rules and facts, rather than adopted
from a language model's proposed answers. Formal--text alignment requires
separate checks, as described in the construction procedures.

\paragraph{Human involvement and consent.}
All manual inspection and annotation review were conducted by the
authors; no external participants or paid annotators were recruited.
The benchmark contains synthetic
cases derived from public rule documents, rather than personal records
collected from individuals. The human-review procedure and instructions
are described in Appendix~\ref{app:error-analysis}.

\paragraph{Model-assisted evaluation and analysis.}
GPT-6-Luna evaluates generated reasoning against selected reference traces.
GPT-6 Astra assesses paired thinking outputs and classifies error mechanisms.
Their outputs undergo structured-record validation and, where applicable,
evidence and consistency checks. These checks do not establish independent
human agreement with the judgments. A human reviewer has reviewed the
64-output error-analysis subset (Appendix~\ref{app:error-analysis}). This non-blind human review covers
the selected subset; it does not establish independent inter-annotator
agreement or adjudication of all annotations. The reported qualitative labels
remain model-assisted assessments, distinct from the formally verified
benchmark answers.

\paragraph{Research and writing assistance.}
AI tools also assist with code development, debugging, experiment scripting,
and manuscript editing. The authors are responsible for the benchmark design,
implementation, interpretation of results, and final manuscript, including
checking AI-assisted material before use.

\end{document}